\documentclass[preprints,article,accept,moreauthors,pdftex,10pt,a4paper]{Definitions/mdpi} 

\firstpage{1} 
\pubvolume{10}
\issuenum{0}
\articlenumber{00}
\pubyear{2018}
\copyrightyear{2018}   \updates{yes} 
\history{Received: 27 July 2018; Accepted: 26 August 2018; Published: date}
 
\usepackage{booktabs} 
\usepackage{multirow}
\usepackage{soul} 
\usepackage{microtype}
\setitemize{parsep=6pt,itemsep=0pt,leftmargin=*,labelsep=5.5mm}
\setenumerate{parsep=6pt,itemsep=0pt,leftmargin=*,labelsep=5.5mm}
\setlist[description]{itemsep=0mm}   

\Title{WeedMap: A Large-Scale Semantic Weed Mapping Framework Using Aerial Multispectral Imaging and Deep Neural Network for Precision Farming}

\Author{Inkyu Sa $^{1,}$*$^{,\dagger}$, Marija Popovi\'{c} $^{1}$, Raghav Khanna $^{1}$, Zetao Chen $^{2}$, Philipp Lottes $^{3}$, Frank~Liebisch~$^{4}$, Juan Nieto $^{1}$, Cyrill Stachniss $^{3}$, Achim Walter $^{4}$ and Roland Siegwart $^{1}$}
\AuthorNames{Authors}

\address{%
$^{1}$ \quad Autonomous Systems Lab., Department of Mechanical and Process Engineering, ETHZ, Zurich~8092,~Switzerland; 
 marija.popovic@mavt.ethz.ch (M.P.); raghav.khanna@mavt.ethz.ch (R.K.); nietoj@ethz.ch (J.N.); rsiegwart@ethz.ch (R.S.)\\
$^{2}$ \quad Vision for Robotics Lab., Department of Mechanical and Process Engineering, ETHZ, Zurich~8092,~Switzerland; 
 chenze@ethz.ch (Z.C.)\\
 $^{3}$ \quad Institute of Geodesy and Geoinformation, University of Bonn, Bonn 53115, Germany; 
philipp.lottes@uni-bonn.de (P.L.); cyrill.stachniss@igg.uni-bonn.de (C.S.)\\
$^{4}$ \quad Crop Science, Department of Environmental Systems Science, ETHZ, Zurich 8092, Switzerland; 
 frank.liebisch@usys.ethz.ch (F.L.); achim.walter@usys.ethz.ch (A.W.)}

\corres{Correspondence: inkyu.sa@mavt.ethz.ch; Tel.: +41-44-632-54-14}

\firstnote{Current address: Leonhardstrasse 21, Building LEE, J, 8092 Zurich, Switzerland.}

\def\fps@figure{htp}
\def\fps@table{htp}

\newcommand{\bi}{\begin{itemize}}
\newcommand{\ei}{\end{itemize}}

\newcommand{\bfig}{\begin{figure}}
\newcommand{\efig}{\end{figure}}

\newcommand{\benum}{\begin{enumerate}}
\newcommand{\eenum}{\end{enumerate}}

\newcommand{\be}{\begin{equation}}
\newcommand{\ee}{\end{equation}}

\newcommand{\ba}{\begin{eqnarray}}
\newcommand{\ea}{\end{eqnarray}}

\newcommand{\etal}{{et al.}}

\newcommand{\unit}[1]{\mbox{$\rm \,#1$}}

\usepackage{color}
\usepackage{graphicx}
\usepackage{setspace}
\usepackage{hyperref}
\usepackage{multirow}
\usepackage{float}
\usepackage{tabulary}
\usepackage{tabularx}
 \usepackage[labelformat=simple]{subfig}

\usepackage{booktabs}
\usepackage{amsmath} 
\usepackage{amssymb}  
\usepackage[flushleft]{threeparttable}
\usepackage{todonotes}
\usepackage{rotating} 
\usepackage{comment}

\definecolor{CommentRed}{rgb}{0.7,0,0}
\definecolor{CommentBlue}{rgb}{0,0,0.7}
\definecolor{CommentDG}{rgb}{0,0.6,0}
\definecolor{CommentPink}{rgb}{1,0.2,0.5}
\definecolor{CommentMagenta}{rgb}{1,0,1}

\newcommand{\iks}[1] {{\color{CommentDG} { iks: \textbf{#1}}}}
\newcommand{\smtxt}[1]{\mbox{\tiny $\text{#1}$}}

\newcommand{\masha}[1] {{\color{CommentPink} { Masha: {#1}}}}

\newcommand{\cyrill}[1]{{\textbf{[Cyrill: {#1}]}}}

\abstract{
The ability to automatically monitor agricultural fields is an important capability in precision farming, enabling steps towards more sustainable agriculture. Precise, high-resolution monitoring is a key prerequisite for targeted intervention and the selective application of agro-chemicals.
{The main goal of} this paper {is developing} a novel {crop/weed} segmentation and mapping framework that processes multispectral images obtained from an unmanned aerial vehicle (UAV) using a deep neural network (DNN).
 Most studies on crop/weed semantic segmentation only consider single images for processing and classification. Images taken by UAVs often cover only a few hundred square meters with either color only or color and near-infrared (NIR) channels. Although a map can be generated by processing single segmented images incrementally, this requires additional complex information fusion techniques which struggle to handle high fidelity maps due to their computational costs and problems in ensuring global consistency. Moreover, computing a single large and accurate vegetation map (e.g., crop/weed) using a DNN is non-trivial due to difficulties arising from: (1) limited ground sample distances (GSDs) in high-altitude datasets, (2) sacrificed resolution resulting from downsampling high-fidelity images, and (3) multispectral image alignment. To address these issues, we adopt a stand sliding window approach that operates on only small portions of multispectral orthomosaic maps (tiles), which are channel-wise aligned and calibrated radiometrically across the entire map. We define the tile size to be the same as that of the DNN input to avoid resolution loss. Compared to our baseline model (i.e., SegNet with 3 channel RGB (red, green, and blue) inputs) yielding an area under the curve (AUC) of [background=0.607, crop=0.681, weed=0.576], our proposed model with 9 input channels achieves [0.839, 0.863, 0.782]. Additionally, we provide an extensive analysis of 20 trained models, both qualitatively and quantitatively, in order to evaluate the effects of varying input channels and tunable network hyperparameters. Furthermore, we release a large sugar beet/weed aerial dataset with expertly guided annotations for further research in the fields of remote sensing, precision agriculture, and agricultural robotics.}


\begin{document}


\section{Introduction}
\label{sec:intro}

Unmanned aerial vehicles (UAVs) are increasingly used as a timely, inexpensive, and agile platform for collecting high-resolution remote sensing data for applications in precision agriculture. With the aid of a global positioning system (GPS) and an inertial navigation system (INS) technology, UAVs can be equipped with commercially available, high-resolution multispectral sensors to collect valuable information for vegetation monitoring. This data can then be processed to guide field management decisions, potentially leading to significant environmental and economical benefits. For~example, the early detection of weed infestations in aerial imagery enables developing site-specific weed management (SSWM) strategies, which can lead to significant herbicide savings, reduced environmental impact, and increased crop yield. 

Enabling UAVs for such applications is an active area of research, relevant for various fields, including remote sensing \cite{De_Castro2018-kx,Kemker2018}, precision agriculture \cite{Zhang2012-km,Lopez-Granados2011-tl,Walter2017-gj}, and agricultural robotics \cite{Detweiler2015,Lottes:2017aa,sa2018ral} and crop science \cite{joalland2018aerial}. In the past years, accelerating developments in data-driven approaches, such as big data and deep neural networks (DNNs) \cite{Carrio2017-vi}, have allowed for unprecedented results in tasks of crop/weed segmentation, plant disease detection, yield estimation, and plant phenotyping \cite{Pound2017-nd}.

\begin{figure}[H]
\centering
\includegraphics[width=\columnwidth]{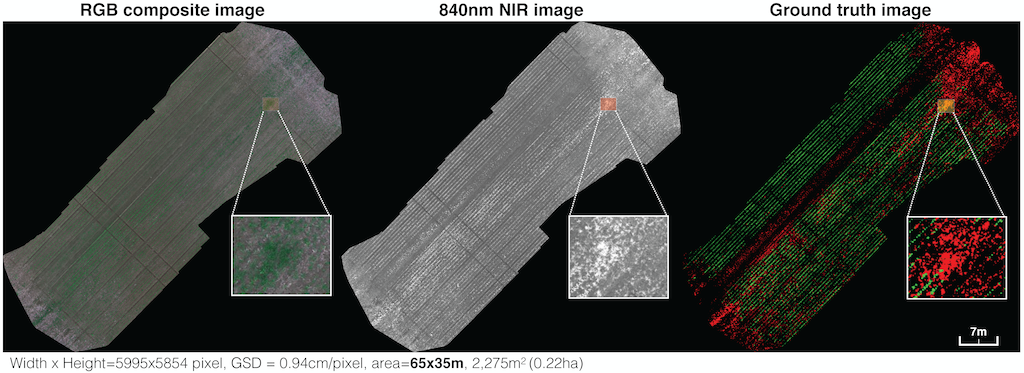}
    \caption{An example of the orthomosaic maps used in this paper. Left, middle and right are RGB (red, green, and blue). 
composite, near-infrared (NIR) and manually labeled ground truth (crop = green, weed = red) images with their zoomed-in views. We present these images in order to provide an intuition of the scale of the sugar beet field and quality of data used in this paper.}
    \label{fig:front}
\end{figure}

However, most practical applications require maps which both cover large areas (on the order~of hectares), while preserving the fine details of the plant distributions. This is a key input for subsequent actions such as weed management. We aim to address this issue by exploiting multispectral orthomosaic maps that are generated by projecting 3D point clouds onto a ground plane, as shown in Figure~\ref{fig:front}.

Utilizing orthomosaic maps in precision agriculture presents several advantages. Firstly, it enables representing crop or field properties of a large farm in a quantitative manner (e.g., a metric scale) by making use of georeferenced images. Secondly, all multispectral orthomosaic maps are precisely aligned, which allows for feeding stacked images to a DNN for subsequent classification. Lastly, global radiometric calibration, i.e., illumination and vignette compensation, is performed over all input images, implying that we can achieve consistent reflectance maps.

There are, of course, also difficulties in using orthomosaic maps. The most prominent one is that the map size may be too large to serve as an input to a standard DNN without losing its resolution {due~to GPU memory limitation}, which may obscure important properties for distinguishing vegetation. Despite recent advances in DNNs, it is still challenging to directly input huge orthomosaic maps to standard classifiers. We address this issue by introducing a sliding window technique that operates on a small part of the orthomosaic before placing it back on the map. The contributions {and aims} of the paper are:
\begin{itemize}
\item The presentation of a complete weed mapping system that operates on large orthomosaic images covering more than 16,500\,m$^2$ (including their labels) and its in-depth performance analysis.
\item The release of unprecedented sugar beet/weed aerial datasets including expertly guided labeled images (hereinafter we refer to the labeled images as ground truth) and corresponding multispectral images \cite{ETH-ASL-Flourish:2018aa}.
\end{itemize}


The remainder of this paper is structured as follows. Section~\ref{sec:related} presents the state of the art in dense semantic segmentation, large-scale vegetation detection using UAVs, and applications of DNNs in precision agriculture. Section~\ref{sec:method} describes our training/testing dataset, and details our orthomosaic generation and processing procedures. We present our experimental results and discuss open challenges/limitations in Sections \ref{sec:experiment} and \ref{sec:discussion}, before concluding in Section \ref{sec:conclusion}.

\section{Related Work}\label{sec:related}

The potentialities of UAV based remote sensing have attracted a lot of interest in high-resolution vegetation mapping scenarios not only due to their environmental impact, but also their economical benefits. In this section, we review the state-of-the-art in plant detection and classification using UAVs, followed~by dense semantic segmentation variants using DNNs and their applications in precision~agriculture. 
\subsection{Vegetation Detection and Classification Using UAVs}

With the aid of rapidly developing fundamental hardware technologies (e.g., sensing, integrated circuit, and battery), software technologies such as machine learning and image processing have played a significant role in remote sensing, agricultural robotics, and precision farming. Among a wide range of agricultural applications, several machine learning techniques have demonstrated remarkable improvements for the task of crop/weed classification in aerial imagery~\cite{Garcia2015be,Guerrero201211149,PerezOrtiz2015533,PerezOrtiz201685}.

Perez-Ortiz \etal~\cite{PerezOrtiz2015533} proposed a weed detection system categorizing image patches into distinct crop, weed, and soil classes based on pixel intensities in multispectral images and geometric information about crop rows. Their work evaluates different machine learning algorithms, achieving overall classification accuracies of 75--87\%. In a later work, the same authors~\cite{PerezOrtiz201685} used a support vector machine classifier for crop/weed detection in RGB images of sunflower and maize fields. They present a method for both inter-row and intra-row weed detection by exploiting the statistics of pixel intensities, textures, shapes and geometrical information.
 
Sandino \etal~\cite{Sandino2018-pa} demonstrated the identification of invasive grasses/vegetation using a decision tree classifier with Red-Green-Blue (RGB) images. Although they employ a rather standard image processing pipeline with traditional handcrafted features, their results show an impressive 95\%+ classification accuracy for different species. Gao \etal~\cite{Gao2018} investigated weed detection by fusing pixel and object-based image analysis (OBIA) for a Random Forest (RF) classifier, combined with a Hough transform algorithm for maize row detection. With an accuracy of 94.5\%, they achieved promising weed mapping results which illustrate the benefit of utilizing prior knowledge of a field set-up (i.e.,~crop~row~detection), in a similar way to our previous work~\cite{Lottes:2017aa}. However, the method was only tested with a small orthomosaic image covering $150$\,\unit{m^2} with a commercial semi-automated OBIA feature extractor. Ana \etal~\cite{De_Castro2018-kx}, on the other hand, proposed an automated RF-OBIA algorithm for early stage intra-, and inter-weed mapping applications by combining Digital Surface Models (plant~height), orthomosaic images, and RF classification for good feature selection. Based on their results, they also developed site-specific prescription maps, achieving herbicide savings of 69--79\% in areas of low infestation. In our previous work by Lottes \etal~\cite{Lottes:2017aa}, we exemplified multi-class crop (sugar beet) and weed classification using an RF classifier on high-resolution aerial RGB images. The~high-resolution imagery enables the algorithm to detect details of crops and weeds leading to the extraction of useful and discriminative features. Through this approach, we achieve a pixel-wise segmentation on the full image resolution with an overall accuracy of 96\% for object detection in a crop vs. weed scenario and up to 86\% in a crop vs. multiple weed species scenario.

Despite the promising results of the aforementioned studies, it is still challenging to characterize agricultural ecosystems sufficiently well. Agro-ecosystems are often multivariate, complex, and~unpredictable using hand-crafted features and conventional machine learning algorithms~\cite{Kamilaris2018-jj}. These difficulties arise largely due to local variations caused by differences in environments, soil, and crop and weed species. Recently, there is a paradigm shift towards data-driven approaches with DNNs that can capture a hierarchical representation of input data. These methods demonstrate unprecedented performance improvements for many tasks, including image classification, object detection, and~semantic segmentation. In the following section, we focus on semantic segmentation techniques and their applications, as these are more applicable to identify plant species than image classification or object detection algorithms in agricultural environments, where objects' boundaries are often unclear and ambiguous.

\subsection{Dense (Pixel-Wise) Semantic Segmentation Using Deep Neural Networks}

The aim of dense semantic segmentation is to generate human-interpretable labels for each pixel in a given image. This fundamental task presents many open challenges. Most existing segmentation approaches rely on convolutional neural networks (CNNs)~\cite{chen2016deeplab,badrinarayanan2017segnet}. Early CNN-based segmentation approaches typically follow a two-stage pipeline, first selecting region proposals and then training a sub-network to infer a pre-defined label for each proposal~\cite{girshick2014rich}. Recently, the semantic segmentation community has shifted to methods using fully Convolutional Neural Networks (FCNNs)~\cite{Long2015-fp}, which can be trained end-to-end and capture rich image information~\cite{chen2016deeplab,dai2015boxsup} because they directly estimate pixel-wise segmentation of the image as a whole. However, due~to sequential max-pooling and down-sampling operations, FCNN-based approaches are usually limited~to low-resolution predictions. Another popular stream in semantic segmentation is the use of an encoder--decoder architecture with skip-connections, e.g., SegNet \cite{badrinarayanan2017segnet}, as a common building block in networks~\cite{li2017arxiv,paszke2016arxiv,ronneberger2015micc}. Our~previous work presented a SegNet-based network, weedNet~\cite{sa2018ral}, which is capable of producing higher-resolution outputs to avoid coarse downsampled predictions. However, weedNet can only perform segmentation~on a single image due to physical Graphics Processing Unit (GPU) memory limitations. Our current work differs in that it can build maps for a much larger field. Although this hardware limitation will~be finally resolved in the future with the development~of parallel computing technologies, to the authors' best knowledge, it is difficult to allocate whole orthomosaic maps (including batches, lossless processing) even on state-of-the art GPU machine memory.

In addition to exploring new neural network architectures, applying data augmentation and utilizing synthetic data are worthwhile options for enhancing the capability of a classifier. These technologies often boost up classifier performance with a small training dataset and stabilize a training phase with a good initialization that can lead to a good neural network convergence. Recently, Kemker \etal~\cite{Kemker2018} presented an impressive study on handling multispectral images with deep learning algorithms. They generated synthetic multispectral images with the corresponding labels for network initialization and evaluated their performance on a new open UAV-based dataset with 18 classes, six~bands, and a GSD of 0.047\unit{m}. Compared to this work, we present a more domain-specific dataset, i.e., it only has three classes but a four-times higher image resolution, a higher number of bands including composite and visual NIR spectral images, and \textbf{1.2} times more data (268\unit{k}/209\unit{k} spectral~pixels).

\subsection{Applications of Deep Neural Networks in Precision Agriculture}

As reviewed by Carrio \etal~\cite{Carrio2017-vi} and Kamilaris \etal~\cite{Kamilaris2018-jj}, the advent of DNNs, especially CNNs, also spurred increasing interest for end-to-end crop/weed classification~\cite{potena2016ias, mortensen2016cigr, milioto2018icra, mccool2017ral, cicco2017iros} to overcome the inflexibility and limitations of traditional handcrafted vision pipelines. In this context, CNNs are applied pixel-wise in a sliding window, seeing only a small patch around a given pixel. Using this principle, Potena \etal~\cite{potena2016ias} presented a multi-step visual system based on RGB and NIR imagery for crop/weed classification using two different CNN architectures. A shallow network performs vegetation detection before a deeper network further discriminates between crops and weeds. They perform a pixel-wise classification followed by a voting scheme to obtain predictions for connected components in the vegetation mask, reporting an average precision of $98$\% if the visual appearance has not changed between the training and testing phases. Inspired by the encoder--decoder network, Milioto \etal~\cite{milioto2018icra} use an architecture which combines normal RGB images with background knowledge encoded in additional input channels. Their work focused on real-time crop/weed classification through a lightweight network architecture. Recently, Lottes \etal~\cite{lottes2018ral} proposed an FCNN-based approach with sequential information for robust crop/weed detection. Their motivation was to integrate information about plant arrangement in order to additionally exploit geometric clues. McCool~\etal~\cite{mccool2017ral} fine-tuned a large CNN \cite{szegedy2016cvpr} for the task at hand and attained efficient processing times by compression of the adapted network using a mixture of small, but fast networks, without sacrificing significant classification accuracy. Another noteworthy approach was presented by\linebreak Mortensen~\etal~\cite{mortensen2016cigr}. They apply a deep CNN for classifying different types of crops to estimate individual biomass amounts. They use RGB images of field plots captured at $3$\,m above the soil and report an overall accuracy of 80\% evaluated on a per-pixel basis. 

Most of the studies mentioned were only capable of processing a single RGB image at a time due~to GPU memory limitations. In contrast, our approach can handle multi-channel inputs to produce more complete weed maps. 
\section{Methodologies}\label{sec:method}
This section presents the data collection procedures, training and testing datasets, and methods~of generating multispectral orthomosaic reflectance maps. Finally, we discuss our dense semantic segmentation framework for vegetation mapping in aerial imagery.

\subsection{Data Collection Procedures}\label{sec:data-collection}

\begin{figure}[H]
\begin{center}
\includegraphics[width=0.8\columnwidth]{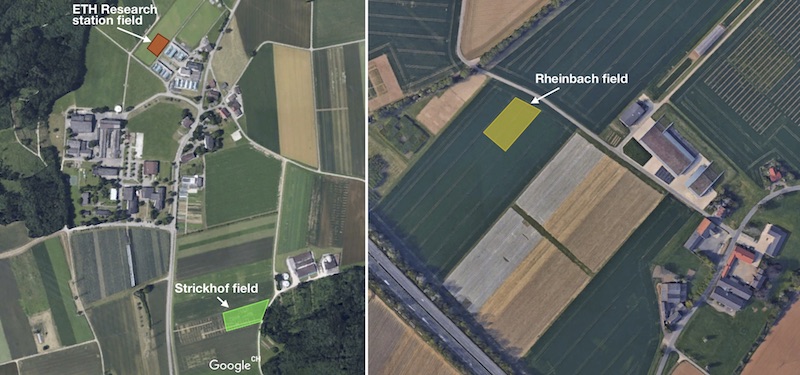}
\end{center}
    \vspace{-3mm}
    \caption{Sugar beet fields where we collected datasets. Two fields in Eschikon are shown on the left, and one field in Rheinbach is on the right.}
    \label{fig:fields}
        \vspace{-3mm}
\end{figure}

Figure~\ref{fig:fields} shows sugar beet fields where we performed dataset collection campaigns. For the experiment at ETH Research station sugar beet (\textit{Beta vulgaris}) of the variety 'Samuela' (KWS Suisse SA, Basel, Switzerland) were sown on 5/April/2017 at 50\unit{cm} row distance, 18\unit{cm} intra row distance. No fertilizater was applied because the soil available Nitrogen was considered sufficient for this short-term trial to monitor early sugar beet growth. The experiment at Strickhof (N-trial field) was sown on 17/March/2017 with the same sugar beet variety and plant density configuration. Fertilizer application was 103$\unit{kg\; N/ha}$ (92$P_{\mbox{\tiny 2}}O_{\mbox{\tiny 5}}$, 360$K_{\mbox{\tiny 2}}O$, 10\unit{Mg}). The fields expressed high weed pressure with large species diversity. Main weeds were \textit{Galinsoga spec., Amaranthus retroflexus, Atriplex spec., Polygonum spec., Gramineae (Echinochloa crus-galli, agropyron} and others.). Minor  weeds were \textit{Convolvulus arvensis, Stellaria  media, Taraxacum spec.} etc. The growth stage of sugar beets ranged from 6 to 8 leaf stage at the moment of data collection campaign (5--18/May/2017) and the sizes of crops and weeds exhibited 8--10 cm and 5--10 cm, respectively.
The sugar beets on the Rheinbach field were sowed on 18/Aug./2017 and their growth stage were about one month at the moment of data collection (18/Sep./2017). The size of crops and weeds were 15--20\unit{cm} and 5--10\unit{cm} respectively. The crops were arranged
at 50\unit{cm} row distance, 20\unit{cm} intra row distance. The field was only treated once during the post-emergence stage of the crops by mechanical weed control action and thus is affected by high weed pressure.

   \unskip

\begin{figure}[H]
\centering
\includegraphics[width=0.8\columnwidth]{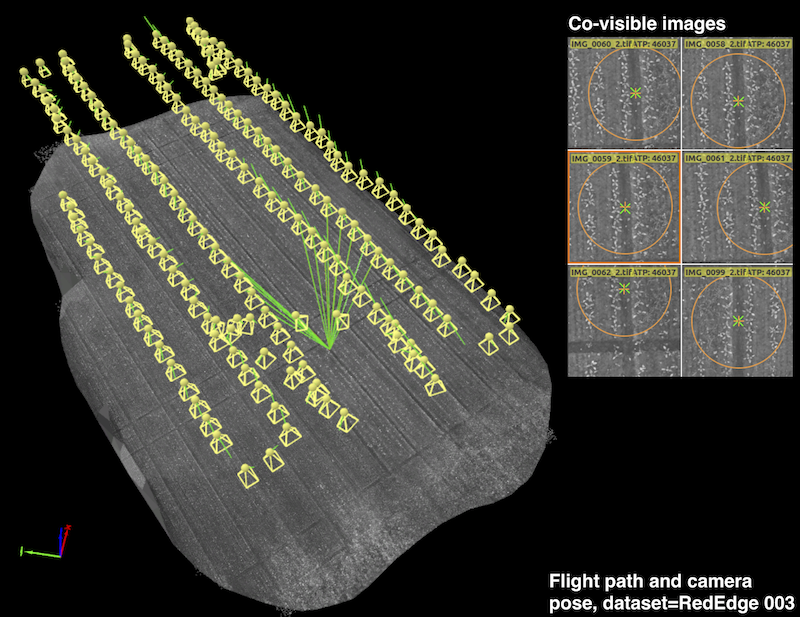}

    \caption{An example UAV trajectory covering a 1300\unit{m^2} sugar beet field (RedEdge-M 002 from Table~\ref{tbl:dataset-detail}). Each yellow frustum indicates the position where an image is taken, and the green lines are rays between a 3D point and their co-visible multiple views. Qualitatively, it can be seen that the 2D feature points from the right subplots are properly extracted and matched for generating a precise orthomosaic map. A similar coverage-type flight path is used for the collection of our datasets.}
    \label{fig:flight-path}
\end{figure}

\begin{figure}[H]
\centering
\includegraphics[width=\columnwidth]{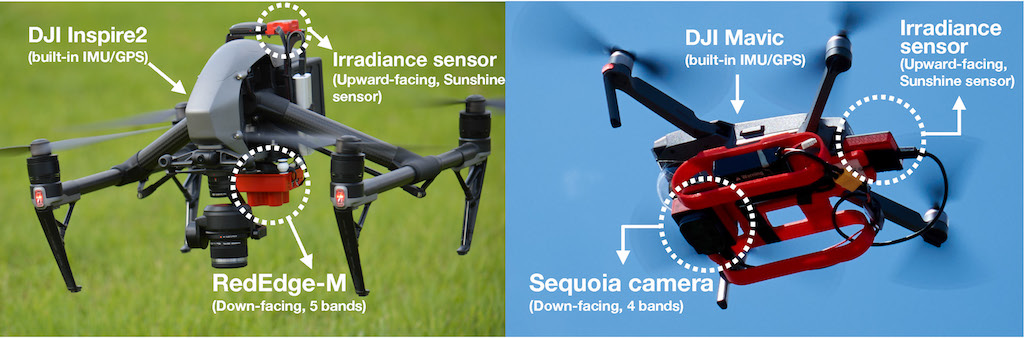}
    \caption{{Multispectral cameras and irradiance (Sunshine) sensors' configuration. Both cameras are facing-down with respect to the drone body and irradiance sensors are facing-up.}}
    \label{fig:sensor-config}
\end{figure}

Figure~\ref{fig:flight-path} illustrates an example dataset we collected (RedEdge-M 002 from Table \ref{tbl:dataset-detail}, Rheinbach, Germany), indicating the flight path and camera poses where multispectral images were registered. Following this procedure, other datasets were recorded at the same altitude and at similar times of day on different sugar beet fields. Table \ref{tbl:data-collection} details our data collection campaigns. Note that an individual aerial platform shown in Figure~\ref{fig:sensor-config} was separately utilized for each sugar beet field. Table \ref{tbl:sensors} elaborates the multispectral sensor specifications, and Tables \ref{tbl:dataset-detail} and \ref{tbl:dataset-overview}   summarize the training and testing datasets for developing our dense semantic segmentation framework in this paper. To assist further research in this area, we make the datasets publicly available \cite{ETH-ASL-Flourish:2018aa}.

\begin{figure}[H]
\centering
\includegraphics[width=\columnwidth]{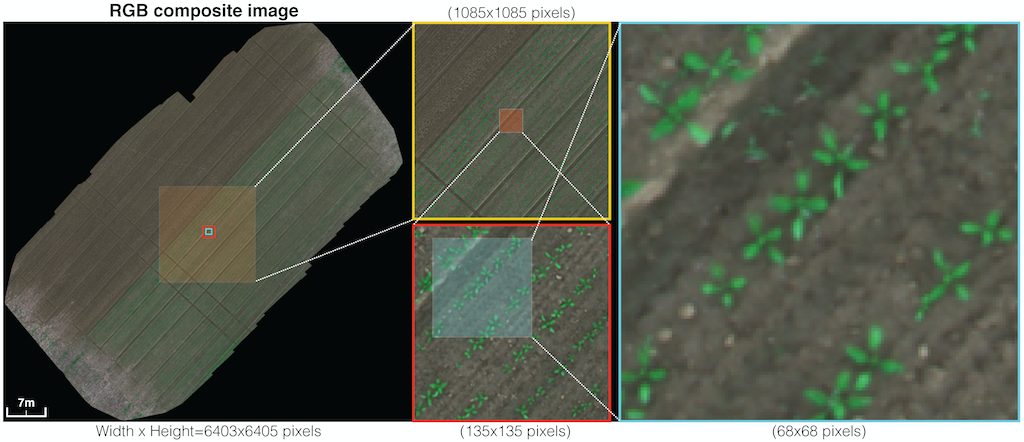}
    \caption{One of the datasets used in this paper. The left image shows the entire orthomosaic map, and the middle and right are subsets of each area at varying zoom levels. The yellow, red, and cyan boxes indicate different areas on the field, corresponding to cropped views. These details clearly provide evidence of the large scale of the farm field and suggest the visual challenges in distinguishing between crops and weeds due to the limited number of pixels and similarities in appearance.}
    \label{fig:redEdge-sample}
\end{figure}

Figure \ref{fig:redEdge-sample} exemplifies the RGB channel of an orthomosaic map generated from data collected with a RedEdge-M camera. The colored boxes in the orthomosaic map indicate areas of different scales on the field, which correspond to varying zoom levels. For example, the cyan box on the far right (68~$\times$~68~pixels) shows a zoomed view of the area within the small cyan box in the orthomosaic map. This figure provides qualitative insight into the high resolution of our map. At the highest zoom level, crop plants are around 15--20 pixels in size and single weeds occupy 5--10 pixels. These clearly demonstrate challenges in crop/weed semantic segmentation due to their small sizes (e.g., 0.05\unit{m} weeds and 0.15\unit{m} crops) and the visual similarities among vegetation. 

  \unskip

\begin{table}[H]
\centering
\caption{Detail of training and testing dataset.}
\label{tbl:dataset-detail}
\scalebox{0.8}[0.95]{\begin{tabular}{ccccccccc}
\toprule
\textbf{Camera}                                                            & \multicolumn{5}{c}{\textbf{RedEdge-M}}                                                            & \multicolumn{3}{c}{\textbf{Sequoia}}                                                        \\ \midrule 
Dataset name                                                                & 000 & 001 & 002 & 003 & \multicolumn{1}{c}{004} & 005 & 006 & 007 \\ \midrule
\begin{tabular}[c]{@{}c@{}}Resolution \\ (col/row) \\ (width/height)\end{tabular}                                                                  & 5995 $\times$ 5854    & 4867 $\times$ 5574    & 6403 $\times$ 6405     & 5470 $\times$ 5995     & \multicolumn{1}{c}{4319 $\times$ 4506}     & 7221 $\times$ 5909                    & 5601 $\times$ 5027                    & 6074 $\times$ 6889                     \\ \midrule
Area covered (ha)                                                                & 0.312	    & 0.1108    & 0.2096	     & 0.1303	     & \multicolumn{1}{c}{0.1307	}     & 0.2519	                    & 0.3316	                    & 0.1785	                     \\ \midrule
GSD (cm)                                                                & 1.04		    & 0.94	    & 
0.96		     & 0.99			     & \multicolumn{1}{c}{ 1.07}     & 0.85		                    & 1.18	                    & 0.83	                     \\ \midrule
\begin{tabular}[c]{@{}c@{}}Tile resolution\\(row/col) pixels\end{tabular}                                                                         & \multicolumn{8}{c}{360/480} \\ \midrule
\# effective tiles                                                          & 107          & 90           & 145           & 94            & \multicolumn{1}{c}{61}            & 210                          & 135                          & 92                            \\ \midrule
\begin{tabular}[c]{@{}c@{}}\# tiles in row/\\\# tiles in col\end{tabular} & 17$\times$13        & 16$\times$11        & 18 $\times$ 14         & 17 $\times$ 12         & \multicolumn{1}{c}{13 $\times$ 9}          & 17 $\times$ 16                        & 14$ \times$ 12                        & 20 $\times$ 13                         \\ \midrule
\begin{tabular}[c]{@{}c@{}}Padding info\\ (row/col) pixels\end{tabular}     & 266/245      & 186/413      & 75/317        & 125/290       & \multicolumn{1}{c}{174/1}         & 211/459                      & 13/159                       & 311/166                       \\ \midrule
Attribute                                                                   & train        & train        & train         & \textbf{test}          & \multicolumn{1}{c}{train}         & \textbf{test}                         & train                        & train                         \\ \midrule
\# channels                                                                 & \multicolumn{5}{c}{5}                                                                           & \multicolumn{3}{c}{4}                                                                       \\ \midrule
Crop                                                                        & \multicolumn{8}{c}{Sugar beet}\\
\bottomrule
\end{tabular}}
\end{table}

As shown in Table \ref{tbl:data-collection}, we collected eight multispectral orthomosaic maps using the sensors specified in Table \ref{tbl:sensors}. The~two~data collection campaigns cover a total area of 1.6554 \unit{ha} (16,554\unit{m^2}). The~two~cameras we used can capture five and four raw image channels, and we compose them to obtain RGB and color-infrared (CIR) images by stacking the R, G, B channels for an RGB image (RedEdge-M) and R, G, and NIR for a CIR image (Sequoia). We also extract the Normalized Difference Vegetation Index (NDVI)~\cite{rouse1973monitoring}, given by a linear correlation, $\text{NDVI}=\frac{(\text{NIR}-\text{R})}{(\text{NIR}+\text{R})}$. These processes {(i.e., color composition for RGB and CIR, and NDVI extraction)} result in 12 and eight channels for the RedEdge-M and Sequoia camera, respectively {(see Table~\ref{tbl:dataset-overview} for the input data composition)}. Although some of channels are redundant (e.g., single G channel and G channel from RGB image), they are processed independently with a subsequent convolution network (e.g., three composed pixels from RGB images are convoluted by a kernel that has a different size as that of a single channel). Therefore, we treat each channel as an image, resulting in a total of \textbf{1.76 billion} pixels composed of 1.39 billion training pixels and 367 million testing pixels (\textbf{10,196} images). To our best knowledge, this is the largest publicly available dataset for a sugar beet field containing multispectral images and their pixel-level ground truth. Table \ref{tbl:dataset-overview} presents an overview of the training and testing folds.

\begin{table}[H]
\centering
\caption{Data collection campaigns summary.}
\label{tbl:data-collection}
\begin{tabular}{ccc}
\toprule
\textbf{Description}                                                             & \textbf{1st Campaign} & \textbf{2nd Campaign} \\  \midrule
Location                                                                         & Eschikon, Switzerland & Rheinbach, Germany    \\ \midrule
Date, Time                                                                        & 5--18 May 2017, around 12:00 p.m.         & 18 September 2017, 9:18--40 a.m.       \\ \midrule
Aerial platform                                                                  & Mavic pro                 & Inspire 2              \\ \midrule
Camera $^a$                                                                           & Sequoia               & RedEdge-M               \\ \midrule
\# Orthomosaic map                                                               & 3                     & 5                     \\ \midrule
\begin{tabular}[c]{@{}c@{}}Training/Testing \\ multispectral images $^b$\end{tabular} & 227/210               & 403/94                \\ \midrule
Crop & \multicolumn{2}{c}{Sugar beet} \\ \midrule
Altitude                                                                         & \multicolumn{2}{c}{10\unit{m}}                  \\ \midrule 
{Cruise speed} $^c$                                                                         & \multicolumn{2}{c}{{4.8}\unit{m/s}} \\\bottomrule                  
\end{tabular}\\\begin{tabular}{@{}c@{}} 
\multicolumn{1}{p{\textwidth -.88in}}{\footnotesize  
 $^a$ See the detail sensor specifications in Table \ref{tbl:sensors};  $^b$ See the detail dataset descriptions in Tables \ref{tbl:dataset-detail} and \ref{tbl:dataset-overview};  $^c$ Front and side overlaps set 80\% and 60\% respectively.}
\end{tabular}  

\end{table}    \unskip

\begin{table}[H]
  \caption{Multispectral camera sensors specifications used in this paper.}
  \label{tbl:sensors}
  \centering
\begin{tabular}{cccc}
\toprule
\textbf{Description}& \textbf{RedEdge-M} & \textbf{Sequoia} & \textbf{Unit}\\  \midrule 
Pixel size & \multicolumn{2}{c}{3.75} & \unit{um} \\ \midrule
Focal length & 5.5 & 3.98 & \unit{mm} \\ \midrule
Resolution (width $\times$ height) & \multicolumn{2}{c}{1280 $\times$ 960} & pixel \\ \midrule
Raw image data bits & 12 & 10 & bit \\ \midrule
Ground Sample Distance (GSD) & 8.2 & 13 & cm/pixel (at 120 m altitude) \\ \midrule
Imager size (width $\times$ height) & \multicolumn{2}{c}{4.8 $\times$ 3.6} & mm \\ \midrule
Field of View (Horizontal, Vertical) & 47.2, 35.4 & 61.9, 48.5 & degree \\ \midrule
Number of spectral bands & 5 & 4 & N/A\\ \midrule
Blue (Center wavelength, bandwidth) & 475, 20 & N/A & nm \\ \midrule
Green & 560, 20 & 550, 40 & nm \\ \midrule
Red & 668, 10 & 660, 40 & nm \\ \midrule
Red Edge  & 717, 10 & 735, 10 & nm \\ \midrule
Near Infrared& 840, 40 & 790, 40 & nm \\\bottomrule
\end{tabular}
\end{table}%

\begin{table}[H]
\centering
\caption{Overview of training and testing dataset.}
\label{tbl:dataset-overview}
\begin{tabular}{ccc}
\toprule
\textbf{Description}              & \textbf{RedEdge-M}   & \textbf{Sequoia}   \\ \midrule 
\# Orthomosaic map              & 5                  & 3                  \\ \midrule
Total surveyed area (ha)                        & 0.8934                 & 0.762                  \\ \midrule
\# channel                        & 12 $^{a}$                & 8 $^{b}$                   \\ \midrule
\begin{tabular}[c]{@{}c@{}}Input image size \\ ({in pixel}, tile size)\end{tabular}                  & \multicolumn{2}{c}{480 $\times$ 360}             \\ \midrule
\multirow{2}{*}{\# training data} & \# images= 403 $\times$ 12 = 4836 $^{c}$        & \# images = 404 $\times$ 8 $^{b}$\;\;=3232         \\
                                  & \# pixel = 835,660,800 & \# pixel = 558,489,600 \\ \midrule
\# testing data                   & 94$ \times$ 12 = 1128         & 125 $\times$ 8 = 1000         \\
                                  & \# pixel = 194,918,400 & \# pixel = 172,800,000 \\ \midrule
Total data                      & \multicolumn{2}{c}{\# image = 10,196, \# pixel = 1,761,868,800}\\ \midrule                              
Altitude                          & \multicolumn{2}{c} {10\unit{m}}\\\bottomrule             
\end{tabular}\\
\begin{tabular}{@{}c@{}} 
\multicolumn{1}{p{\textwidth -.88in}}{\footnotesize 
  $^{a}$ 12 channels of RedEdge-M data consists of R(1), Red edge(1), G(1), B(1), RGB(3), CIR(3), NDVI(1), and NIR(1). The number in parentheses indicate the number of channel; $^{b}$ 8 channels Sequoia data consists of R(1), Red edge(1), G(1), CIR(3), NDVI(1), and NIR(1); $^{c}$ Each channel is treated as an image.}
\end{tabular}
\end{table}

\subsection{Training and Testing Datasets}

The input image size refers to the resolution of data received by our DNN. Since most CNNs downscale input data due to the difficulties associated with memory management in GPUs, we~define the input image size to be the same as that of the input data. This way, we avoid the down-sizing operation, which significantly degrades classification performance by discarding crucial visual information for distinguishing crop and weeds. 
Note that \textbf{tile} implies that a portion of the region in an image has the same size as that of the input image. We crop multiple tiles from an orthomosaic map by sliding a window over it until the entire map is covered.

Table \ref{tbl:dataset-detail} presents further details regarding our datasets. The Ground Sample Distance (GSD) indicates the distance between two pixel centers when projecting them on the ground given a sensor, pixel size, image resolution, altitude, and camera focal length, as defined by its field of view (FoV). Given the camera specification and flight altitude, we achieved a GSD of around 1\unit{cm}. This is in line with the sizes of crops (15--20 pixels) and weeds (5--10 pixels) depicted in Figure~\ref{fig:redEdge-sample}. 

\begin{figure}[H]
\centering
\includegraphics[width=\columnwidth]{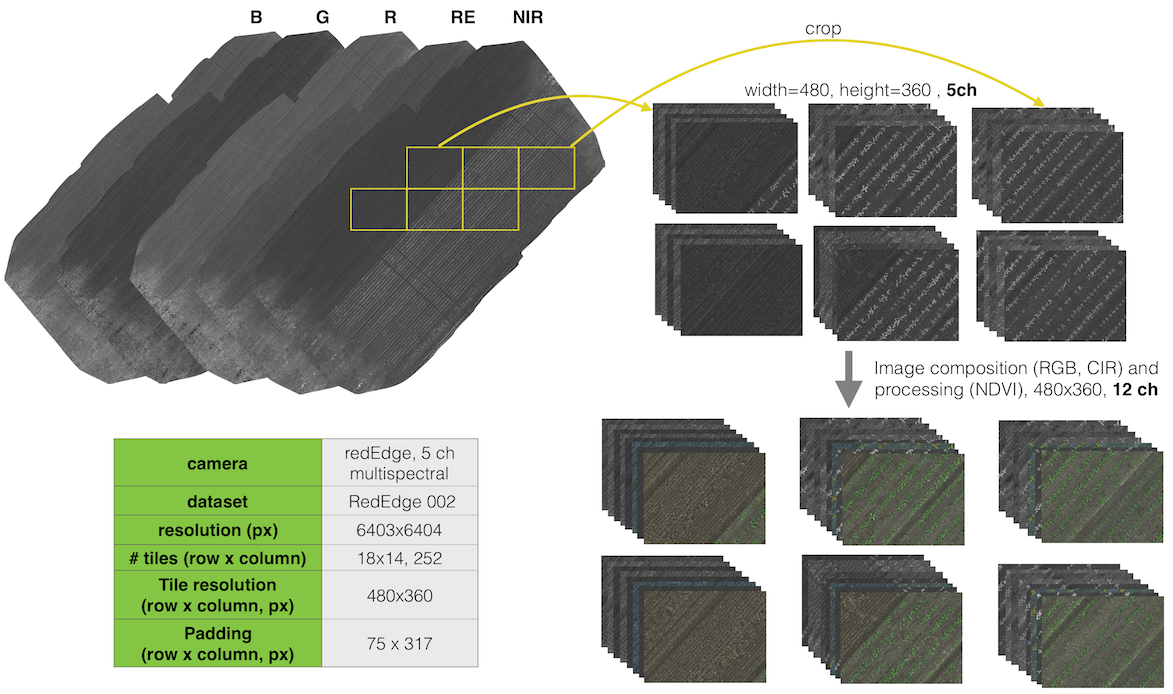}
    \caption{An illustration of tiling from aligned orthomosaic maps. Multispectral images of fixed size (\textbf{top right}) are cropped from aligned orthomosaic maps (\textbf{top left}). Image composition and preprocessing are then performed for generating RGB, CIR, and NDVI respectively. This yields 12 composited tile channels that are input into a Deep Neural Network (DNN).}
    \label{fig:tiling}
\end{figure}

The number of effective tiles is the number of images actually containing any valid pixel values other than all black pixels. This occurs because orthomosaic maps are diagonally aligned such that the tiles from the most upper left or bottom right corners are entirely black images. The~number~of tiles in row/col indicates how many tiles (i.e., 480 $\times$ 360 images) are~composed~in a row and column, respectively. Padding information denotes the number of additional black pixels in rows and columns~to match the size of the orthomosaic map with a given tile size. For example, the~RedEdge-M~\texttt{000} dataset has a size of 5995 $\times$ 5854 for width (column) and height (row), with 245 and 266 pixels appended to the column and row, respectively. This results in a 6240 $\times$ 6210 orthomosaic map consisting of 17 row tiles (17 $\times$ 360 pixels) and 13 column tiles (13 $\times$ 480 pixels). This~information is used when generating a segmented orthomosaic map and its corresponding ground truth map from the tiles. For better visualization, we also present the tiling preprocessing method for the RedEdge-M~002 dataset in Figure~\ref{fig:tiling}. The last property, attribute, shows whether the datasets were utilized for training or testing.

\subsection{Orthomosaic Reflectance Maps}\label{reflectance-map}
\begin{figure}[H]
\centering
\subfloat[]{\includegraphics[height=0.49\columnwidth]{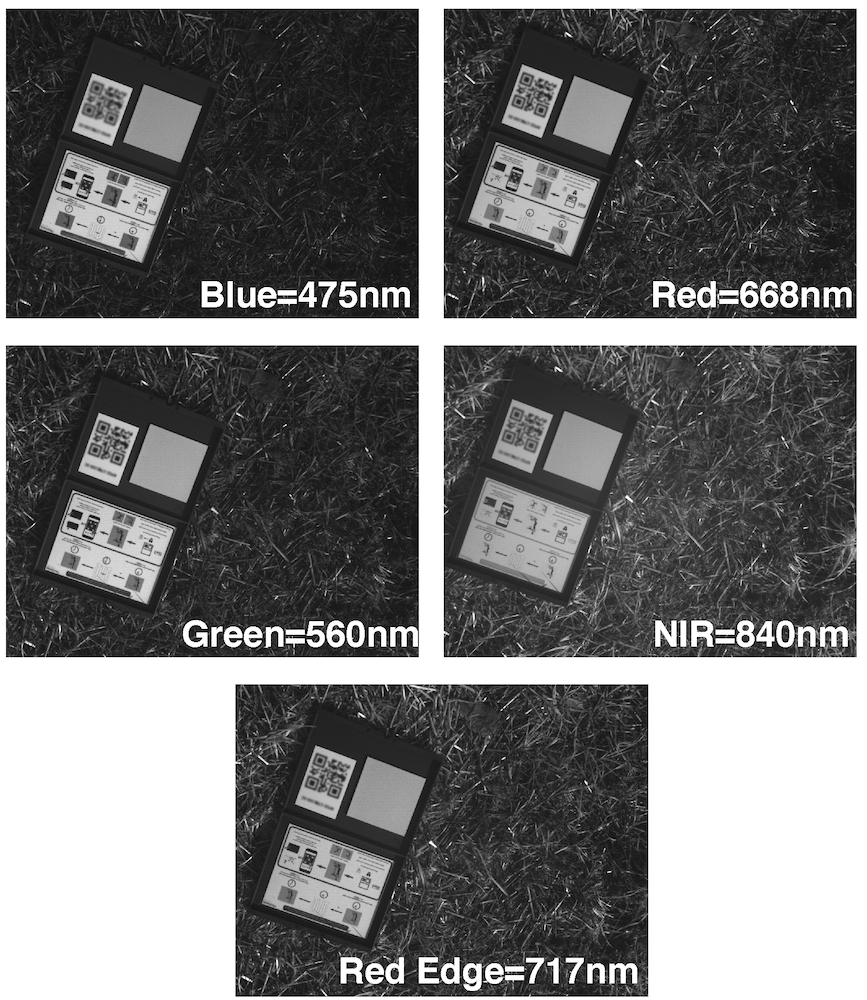}}
\subfloat[]{\includegraphics[height=0.4\columnwidth]{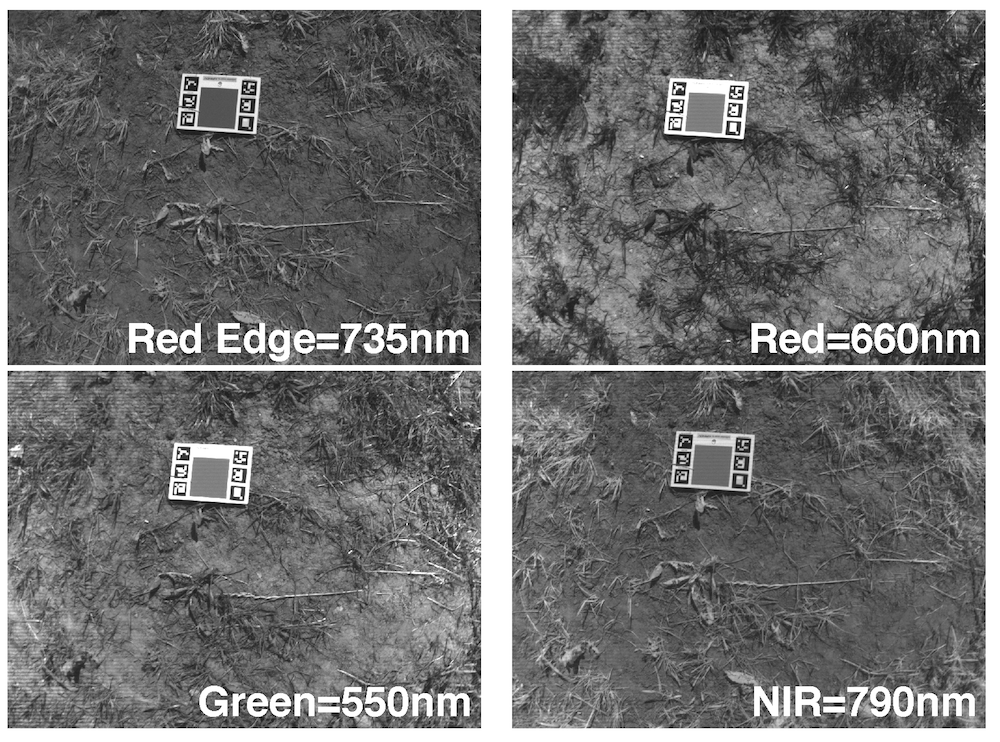}}\newline
	\caption{(\textbf{a}) RedEdge-M radiometric calibration pattern (\textbf{b}) Sequoia calibration pattern for all four~bands.}
	\label{fig:CRP}
\end{figure}

The output from the orthomosaic tool is the reflectance of each band, $r(i,j)$,
\begin{align}
r(i,j) = p(i,j)\cdot f_k,
\label{eq:reflectance}
\end{align}
where $p(i,j)$ is the value of the pixel located in the $i$th row and $j$th column, ordered from top to bottom and left to right in the image, and the top left most pixel is indexed by $i$ = 0, and $j$ = 0. $f_k$ is the reflectance calibration factor of band $k$, which can be expressed by~\cite{MicaSense:2018aa}:
\begin{align}
f_k = \frac{\rho_k}{\text{avg(}L_k\text{)}}\text{,}
\label{eq:calibFactor}
\end{align}
where $\rho_k$ is the average reflectance of the calibrated reflectance panel (CRP) for the $k$th band (Figure~\ref{fig:CRP}), as~provided by the manufacturer, $L_k$ is the radiance for the pixels inside the CRP of the $k$th band. The radiance (unit of watt per steradian per square metre per nanometer, $W/m^2/sr/nm$) of a pixel, $L(i,j)$, can be written as:
\begin{align}
L(i,j) = V(i,j)\cdot \frac{k_{a_1}}{k_{\smtxt{gain}}}\cdot \frac{\bar{p}(i,j)-\bar{p}_{\smtxt{BL}}}{k_{\smtxt{expo}}+k_{a_2}\cdot j-k_{a_3}\cdot k_{\smtxt{expo}}\cdot j}\text{,}
\label{eq:radiance}
\end{align}
where $k_{a_{1:3}}$ are the radiometric calibration coefficients, $k_{\smtxt{expo}}$ is the camera exposure time, $k_{\smtxt{gain}}$ is the sensor gain, and $\bar{p}=p(i,j)/2^n$ and $\bar{p}_{\smtxt{BL}}$ denote the normalized pixel and black level, respectively. $n$ is the number of bits in the image (e.g., $n$ = 12 or 16 bits). $V(i,j)$ is the 5th order radial vignette model, expressed as:
\begin{align}
V(i,j)=&\frac{p(i,j)}{C}, \text{where } C=1+\sum_{i=0}^{5}q_i\cdot r^{i+1}\text{,}\\
r=&\sqrt{(i-c_i)^2+(j-c_j)^2}\text{,}
\label{eq:vignette}
\end{align}
where $q_i$ is vignette coefficient, and $r$ is the distance of the pixel located at ($i,j$) from the vignette center ($c_i, c_j$).
This radiometric calibration procedure is critical to generate a consistent orthomosaic output. We~capture two sets of calibration images (before/after) for each data collection campaign, as~shown~in Figure~\ref{fig:CRP}. To obtain high-quality, uniform orthomosaics (i.e., absolute reflectance), it is important to apply corrections for various lighting conditions such as overcast skies and partial cloud coverage. To~correct~for this aspect, we utilize sunlight sensors measuring the sun's orientation and sun irradiance, as shown in Figure~\ref{fig:diagram}.

\subsubsection{Orthomosaic Map Generation}
Creating orthomosaic images differs to ordinary image stitching as it transforms perspectives to the nadir direction (a top-down view orthogonal to a horizontal plane) and, more importantly, performs true-to-scale operations in which an image pixel corresponds to a metric unit~\cite{fsr_hinzmann_2017, oettershagenrobotic}. This~procedure consists of three key steps: (1) initial processing, (2) point densification, and (3) DSM and orthomosaic generation. Step (1) performs keypoints extraction and matching across the input images. A global bundle adjustment method~\cite{snavely2006photo} optimizes the camera parameters, including the intrinsic (distortions, focal length, and principle points) and extrinsic (camera pose) parameters, and triangulated sparse 3D points (structures). Geolocation data such as GPS or ground control points (GCP) are utilized~to recover the scale. In Step (2), the 3D points are then densified and filtered~\cite{furukawa2010accurate}. Finally, Step (3) back-projects the 3D points on a plane to produce 2D orthomosaic images with a nadir view.

\newpage

\begin{figure}[H]
\centering
\includegraphics[width=.9\columnwidth]{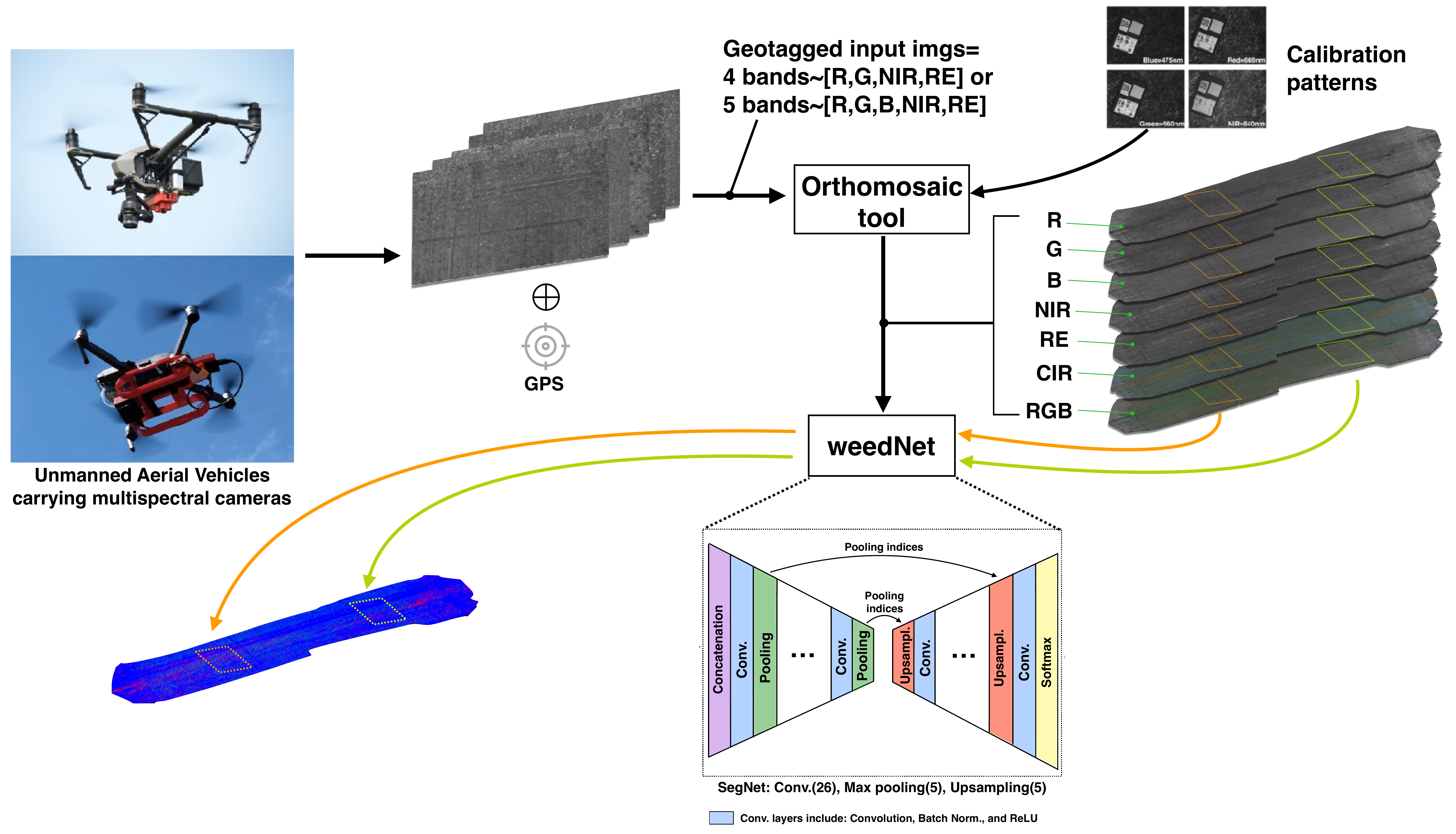}
    \caption{Our overall processing pipeline. GPS tagged multispectral images are first collected by multiple UAVs and then passed to an orthomosaic tool with images for radiometric calibration. Multi-channel and aligned orthomosaic images are then tiled into a small portion (480 $\times$ 360 \unit{pixels}, as~indicated by the orange and green boxes) for subsequent segmentation with a DNN. This operation is repeated in a sliding window manner until the entire orthomosaic map is covered.}
    \label{fig:diagram}
\end{figure}

Since these orthomosaic images are true to scale (metric), all bands are correctly \texttt{aligned}. This~enables using tiled multispectral images as inputs to the subsequent dense semantic segmentation framework presented in Section~\ref{densenetwork}. Figure~\ref{fig:diagram} illustrates the entire pipeline implemented in this paper. First, GPS tagged raw multispectral images (five and four channels) are recorded by using two commercial quadrotor UAV platforms which fly over sugar beet fields. Predefined coverage paths at 10\unit{m} with 80\% side and front overlap between consecutive images are passed to the flight controller. The orthomosaic tool \cite{Pix4D:2018aa} is exploited to generate statistics (e.g., GSD, area coverage, and map uncertainties) and orthomosaic reflectance maps with the calibration patterns presented in Section~\ref{reflectance-map}. Based on these reflectance maps, we compose orthomosaic maps, such as RGB, CIR, and NDVI, and tile them as the exact input size for the subsequent dense semantic framework, weedNet~\cite{sa2018ral}, to avoid downscaling. The predictive output containing per-pixel probabilities for each class has the same size as that of the input, and is returned to the original tile location in the orthomosaic map. This methodology is repeated for each tile to ultimately create a large-scale weed map of the target area.

\subsection{Dense Semantic Segmentation Framework}\label{densenetwork}
In this section, we summarize the dense semantic segmentation framework introduced in our previous work \cite{sa2018ral}, highlighting only key differences with respect to the original implementation. Although our approach relies on a modified version of the SegNet architecture \cite{badrinarayanan2017segnet}, it can be easily replaced with any state-of-the-art dense segmentation tool, such as \cite{Romera2018-jq,paszke2016arxiv,milioto2018icra}.

\subsubsection{Network Architecture}\label{sec:network}
We use the original SegNet architecture in our DNN, i.e., an encoding part with VGG16 layers~\cite{simonyan2014very} in the first half which drops the last two fully-connected layers, followed by upsampling layers for each counterpart in the corresponding encoder layer in the second half. As introduced in~\cite{Garcia-Garcia2017-dv}, SegNet exploits max-pooling indices from the corresponding encoder layer to perform faster upsampling compared to an FCN~\cite{Long2015-fp}.

Our modifications are two-fold. Firstly, the frequency of appearance (FoA) for each class is adapted based on our training dataset for better class balancing~\cite{eigen2015predicting}. This is used to weigh each class inside the neural network loss function and requires careful tuning. A class weight can be written as:
    \vspace{-0mm}
\begin{align}
w_{\mbox{\tiny c}}=&\frac{\widetilde{FoA(c)}}{FoA(c)}\text{,}\\
FoA(c)=&\frac{I_{\mbox{\tiny c}}^{\mbox{\tiny Total}}}{I_{\mbox{\tiny 
c}}^{\mbox{\tiny j}}}\text{,}
\label{eq:pfh}
\end{align}
where $\widetilde{FoA(c)}$ is the median of $FoA(c)$, $I_{\mbox{\tiny 
c}}^{\mbox{\tiny Total}}$ is the total number of pixels in class $c$, and 
$I_{\mbox{\tiny c}}^{\mbox{\tiny j}}$ is the~number~of pixels in the $j$th image 
where class $c$ appears, with $j\in{\{1,2,3, \ldots, N\}}$ as the image sequence number ($N$ indicates the total number of images).

In agricultural context, the \texttt{weed} class usually appears less frequently than \texttt{crop}, thus having a comparatively lower FoA. If a false-positive or false-negative is detected in weed classification, i.e., a~pixel~is incorrectly classified as \texttt{weed}, then the classifier is penalized more for it in comparison to the other classes. We acknowledge that this argument is difficult to generalize to all sugar beet fields, which likely have very different crop/weed ratios compared to our dataset. {More specifically}, the RedEdge-M dataset has $w_{\mbox{\tiny c}}=[0.0638, 1.0, 1.6817]$ for [background, crop, weed] (hereinafter background referred to as \texttt{bg}) classes {with $\widetilde{FoA(c)}$=0.0586 and $FoA(c)$=[0.9304, 0.0586, 0.0356]. This means that 93\%~of pixels in the dataset belong to background class, 5.86\% is crop, and 3.56\% is weed. Sequoia dataset's $w_{\mbox{\tiny c}}$ is [0.0273, 1.0, 4.3802] with $\widetilde{FoA(c)}$=0.0265 and $FoA(c)$=[0.9732, 0.0265, 0.0060]}.

Secondly, we implemented a simple input/output layer that reads images and outputs them~to the subsequent concatenation layer. This allows us to feed any number of input channels of an image~to the network, which contributes additional information for the classification task~\cite{Khanna:2017aa}.

\section{Experimental Results}\label{sec:experiment}
In this section, we present our experimental setup, followed by our quantitative and qualitative results for crop/weed segmentation. The purpose of these experiments is to investigate the performance of our classifier with datasets varying in input channels and network hyperparameters.

\subsection{Experimental Setup}
As shown in Table~\ref{tbl:dataset-detail}, we have eight multispectral orthomosaic maps with their corresponding manually annotated ground truth labels. We consider three classes, \texttt{bg}, \texttt{crop}, and \texttt{weed}, identified numerically by [0, 1, 2]. In all figures in this paper, they are colorized as [\color{blue} bg, \color{CommentDG} crop, \color{red} weed\color{black}]. 

We used datasets [000, 001, 002, 004] for RedEdge-M (5 channel) training and 003 for testing. Similarly, datasets [006, 007] are used for Sequoia (4 channel) training and 005 for testing. Note that we could not combine all sets for training and testing mainly because their multispectral bands are not matched. Even though some bands of the two cameras overlap (e.g., green, red, red-edge, and NIR), the center wavelength and bandwidth, and the sensor sensitivities vary.

For all model training and experimentation, we used the following hyperparameters: learning rate = 0.001, max. iterations = 40,000, momentum = 9.9, weight decay = 0.0005, and gamma = 1.0. We~perform two-fold data augmentation, i.e., the input images are horizontally mirrored.

\subsection{Performance Evaluation Metric}

For the performance evaluation, we use the area under the curve (AUC) of a precision-recall curve~\cite{Boyd2013-pb}, given by:
\begin{align}
\text{precision}_{\mbox{\tiny c}}=\frac{TP_{\mbox{\tiny c}}}{TP_{\mbox{\tiny c}}+FP_{\mbox{\tiny c}}}, \;\;\;
\text{recall}_{\mbox{\tiny c}}=\frac{TP_{\mbox{\tiny c}}}{TP_{\mbox{\tiny c}}+FN_{\mbox{\tiny c}}},
\label{eq:pfh}
\end{align}
where $TP_{\mbox{\tiny c}}$, $TF_{\mbox{\tiny c}}$, $FP_{\mbox{\tiny c}}$, $FN_{\mbox{\tiny c}}$ are the four fundamental numbers, i.e., the numbers of true positive, true negative, false positive, and false negative classifications for class $c$. 
The outputs of the network (480~$\times$~360~$\times$ 3) are the probabilities of each pixel belonging to each defined class. For example, the elements [1:480, 1:360, 2]~\cite{MATLAB:image-coordi} correspond to pixel-wise probabilities for being \texttt{crop}.
To calculate $TP_{\mbox{\tiny c}}$, $TF_{\mbox{\tiny c}}$, $FP_{\mbox{\tiny c}}$, $FN_{\mbox{\tiny c}}$, these probabilities should be converted into binary values given a threshold. Since it is often difficult to find the optimal threshold for each class, we exploit \texttt{perfcurve}~\cite{MATLAB:perfcurve} that incrementally varies thresholds from 0 to 1 and computes $\text{precision}_{\mbox{\tiny c}}$, $\text{recall}_{\mbox{\tiny c}}$, and the corresponding AUC. We believe that computing AUC over the probabilistic output can reflect classification performance better than other metrics \cite{Csurka2013-id}.

For tasks of dense semantic segmentation, there are many performance evaluation metrics \cite{Garcia-Garcia2017-dv} such as Pixel Accuracy (PA), Mean Pixel Accuracy (MPA), Mean Intersection over Union (MIoU), and Frequency Weighted Intersection over Union (FWIoU). All these metrics either rely on specific thresholds or assign the label with maximum probability among all classes in order to compare individual predictions to ground truth. For instance, a given pixel with a ground truth label of 2 and predictive output label of 3 can be considered a false positive for class 2. However, if a pixel receives probabilistic classifications of 40\%, 40\%, and 20\% for classes 1, 2, and 3, respectively, it~may~be inappropriate to apply a threshold or choose the maximum probability to determine its predictive~output.
\subsection{Results Summary}
Table~\ref{tbl:resultSummary} displays the dense segmentation results using 20 different models, varying in the number~of input channels, batch size, class balance flag, and AUC of each class. Model numbers 1--13 and 14--20 denote the RedEdge-M and Sequoia datasets, respectively. Bold font is used to designate the best scores. Figures~\ref{fig:rededge-AUC-bars} and \ref{fig:rededge-AUC-curves} show the AUC scores of each class for the RedEdge-M dataset models and their corresponding AUC curves. Analogously, Figures~\ref{fig:sequoia-AUC-bars} and \ref{fig:sequoia-AUC-curves} depict the AUC scores for the Sequoia dataset models and their AUC curves. The following sections present a detailed discussion and analysis of these results.

\setlength{\tabcolsep}{0.19em} 
{\renewcommand{\arraystretch}{0.8} 
\begin{table}[H]
\centering
\caption{Performance evaluation summary for the two cameras with varying input channels.}
\label{tbl:resultSummary}
\begin{tabular}{cccccccc}
\toprule
                   \multicolumn{5}{c}{\textbf{RedEdge-M}}                                                                                    & \multicolumn{3}{c}{\textbf{AUC} $^b$}            \\ \midrule 
\textbf{\# Model} & \textbf{\# Channels}                                            & \textbf{Used Channel} $^a$            & \textbf{\# batches} & \multicolumn{1}{c}{\textbf{Cls bal.}} & \textbf{Bg} & \textbf{Crop} & \textbf{Weed} \\ \midrule
1                 & 12                                                              & B, CIR, G, NDVI, NIR, R, RE, RGB & 6                   & \multicolumn{1}{c}{Yes}                    & 0.816       & 0.856         & 0.744        \\ \midrule
2                 & 12                                                              & B, CIR, G, NDVI, NIR, R, RE, RGB & 4                   & \multicolumn{1}{c}{Yes}                    & 0.798       & 0.814         & 0.717         \\ \midrule
3                 & 12                                                              & B, CIR, G, NDVI, NIR, R, RE, RGB & 6                   & \multicolumn{1}{c}{No}                     & 0.814       & 0.849         & 0.742         \\ \midrule
4                 & 11        & \begin{tabular}[c]{@{}c@{}} B, CIR, G, NIR, R, RE, RGB \\ (NDVI drop)\end{tabular}       & 6                   & \multicolumn{1}{c}{Yes}                    & 0.575       & 0.618         & 0.545         \\ \midrule
5                 & 9          & \begin{tabular}[c]{@{}c@{}}B, CIR, G, NDVI, NIR, R, RE\\ (RGB drop)\end{tabular}      & 5                   & \multicolumn{1}{c}{Yes}                    & \textbf{0.839}       & \textbf{0.863}         & \textbf{0.782}         \\ \midrule
6                 &9          &  \begin{tabular}[c]{@{}c@{}}B, G, NDVI, NIR, R, RE, RGB\\ (CIR drop)\end{tabular}      & 5                   & \multicolumn{1}{c}{Yes}                    & 0.808       & 0.851         & 0.734         \\ \midrule
7                 & 8 & \begin{tabular}[c]{@{}c@{}}B, G, NIR, R, RE, RGB\\ (CIR and NDVI drop)\end{tabular}            & 5                   & \multicolumn{1}{c}{Yes}                    & 0.578       & 0.677         & 0.482         \\ \midrule
8                 & 6                                                               & G, NIR, R, RGB                   & 5                   & \multicolumn{1}{c}{Yes}                    & 0.603       & 0.672         & 0.576         \\ \midrule
9                 & 4                                                               & NIR, RGB                         & 5                   & \multicolumn{1}{c}{Yes}                    & 0.607       & 0.680         & 0.594         \\ \midrule
10                & 3   & \begin{tabular}[c]{@{}c@{}}RGB\\ (SegNet baseline)\end{tabular}                              & 5                   & \multicolumn{1}{c}{Yes}                    & 0.607       & 0.681         & 0.576         \\ \bottomrule
\end{tabular}

\end{table}  \unskip

\begin{table}[H]\ContinuedFloat
\centering \small
\caption{{\em Cont.}} \label{tbl:resultSummary}
\begin{tabular}{cccccccc}
\toprule
                   \multicolumn{5}{c}{\textbf{RedEdge-M}}                                                                                    & \multicolumn{3}{c}{\textbf{AUC}}            \\ \midrule 
\textbf{\# Model} & \textbf{\# Channels}                                            & \textbf{Used Channel}            & \textbf{\# batches} & \multicolumn{1}{c}{\textbf{Cls bal.}} & \textbf{Bg} & \textbf{Crop} & \textbf{Weed} \\ \midrule

11                & 3  & \begin{tabular}[c]{@{}c@{}}B, G, R\\ (Splitted channel)\end{tabular}                          & 5                   & \multicolumn{1}{c}{Yes}                    & 0.602       & 0.684         & 0.602         \\ \midrule
12                & 1                                                               & NDVI                             & 5                   & \multicolumn{1}{c}{Yes}                    & 0.820       & 0.858         & 0.757         \\ \midrule
13                & 1                                                               & NIR                              & 5                   & \multicolumn{1}{c}{Yes}                    & 0.566       & 0.508         & 0.512         \\ \midrule
                  \multicolumn{5}{c}{\textbf{Sequoia}}                                                                             & \multicolumn{3}{c}{\textbf{AUC}}            \\ \midrule 
14                & 8                                                               & CIR, G, NDVI, NIR, R, RE         & 6                   & \multicolumn{1}{c}{Yes}                    & 0.733       & 0.735         & 0.615         \\ \midrule
15                & 8                                                               & CIR, G, NDVI, NIR, R, RE         & 6                   & \multicolumn{1}{c}{No}                     & 0.929       & 0.928         & 0.630         \\ \midrule
16                & 5                                                               & G, NDVI, NIR, R, RE              & 5                   & \multicolumn{1}{c}{Yes}                    & \textbf{0.951}       & \textbf{0.957}         & 0.621         \\ \midrule
17                & 5                                                               & G, NDVI, NIR, R, RE              & 6                   & \multicolumn{1}{c}{Yes}                    & 0.923       & 0.924         & 0.550         \\ \midrule
18                & 3                                                               & G, NIR, R                        & 5                   & \multicolumn{1}{c}{No}                     & 0.901       & 0.901         & 0.576         \\ \midrule
19                & 3                                                               & CIR                              & 5                   & \multicolumn{1}{c}{No}                     & 0.883       & 0.88          & 0.641         \\ \midrule
20                & 1                                                               & NDVI                             & 5                   & \multicolumn{1}{c}{Yes}                    & 0.873       & 0.873         & \textbf{0.702}\\\bottomrule     
\end{tabular}

\begin{tabular}{@{}c@{}} 
\multicolumn{1}{p{\textwidth -.88in}}{\footnotesize  
 $^a$ R, G, B, RE, NIR indicate red, green, blue, red edge, and near-infrared channel respectively. $^b$ AUC is Area Under the Curve.}
\end{tabular}
\end{table} \unskip

\begin{figure}[H]
\centering
\includegraphics[width=\columnwidth]{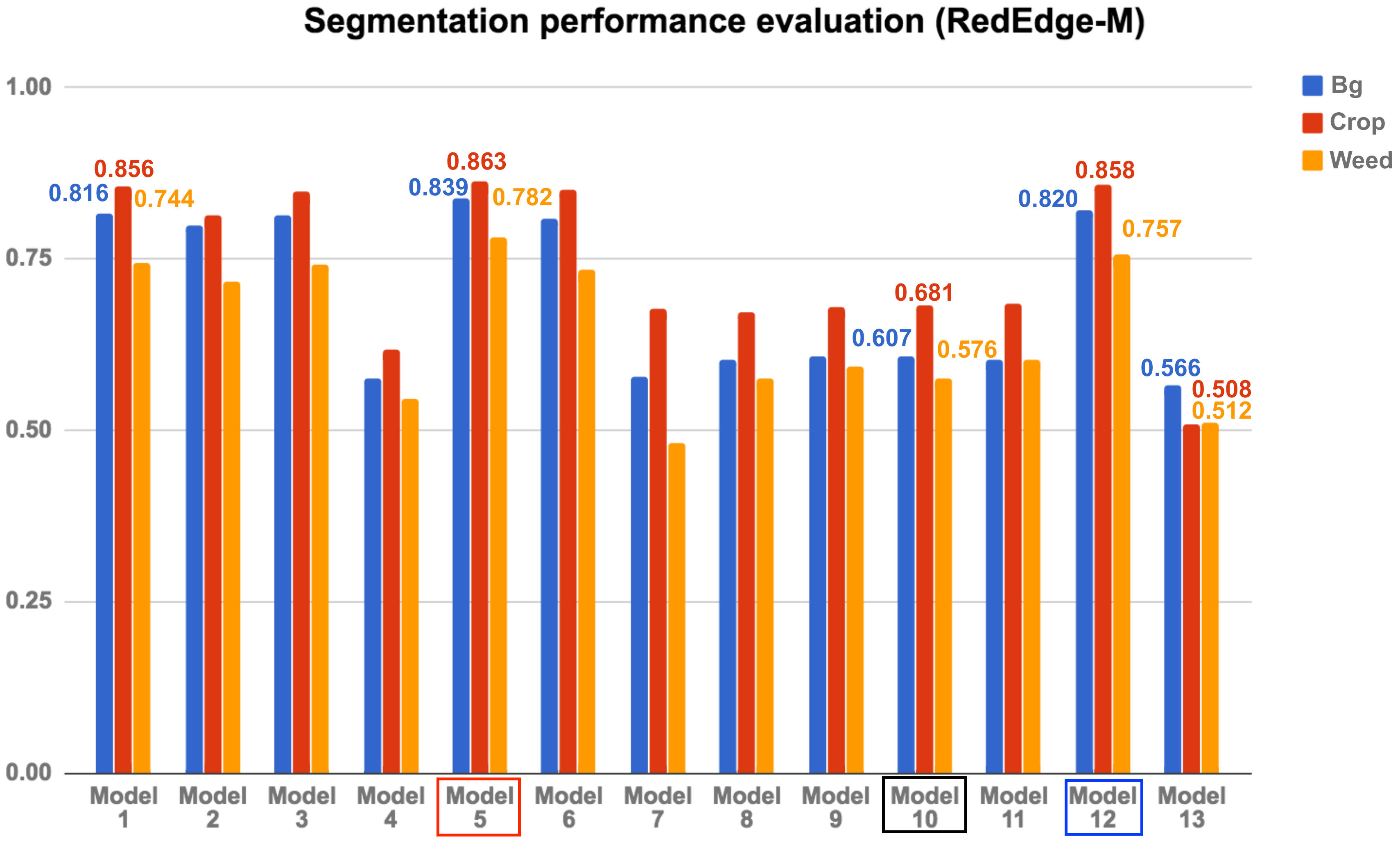}
    \caption{Quantitative evaluation of the segmentation using area under the curve (AUC) of the RedEdge-M dataset. The red box indicates the best model, the black one is our baseline model with only RGB image input, and the blue box is a model with only one NDVI image input.}
    \label{fig:rededge-AUC-bars}
\end{figure}

\subsubsection{Quantitative Results for the RedEdge-M Dataset}
Our initial hypothesis postulated that performance would improve by adding more training data. This argument is generally true, as made evident by Model~10 (our baseline model, the vanilla SegNet with RGB input) and Model~1, but not always; Model 1 and Model 5 present a counter-example. Model~1 makes use of all available input data, but slightly underperforms in comparison to Model 5, which performs best with nine input channels. Although the error margins are small ($<$2\%), this can happen if the RGB channel introduces features that deviate from other features extracted from other channels. As a result, this yields an ambiguity in distinguishing classes and degrades the performance.
\def\picWidth{0.25}
\begin{figure}[H]
\centering
\subfloat[]{\includegraphics[width=\picWidth\columnwidth]{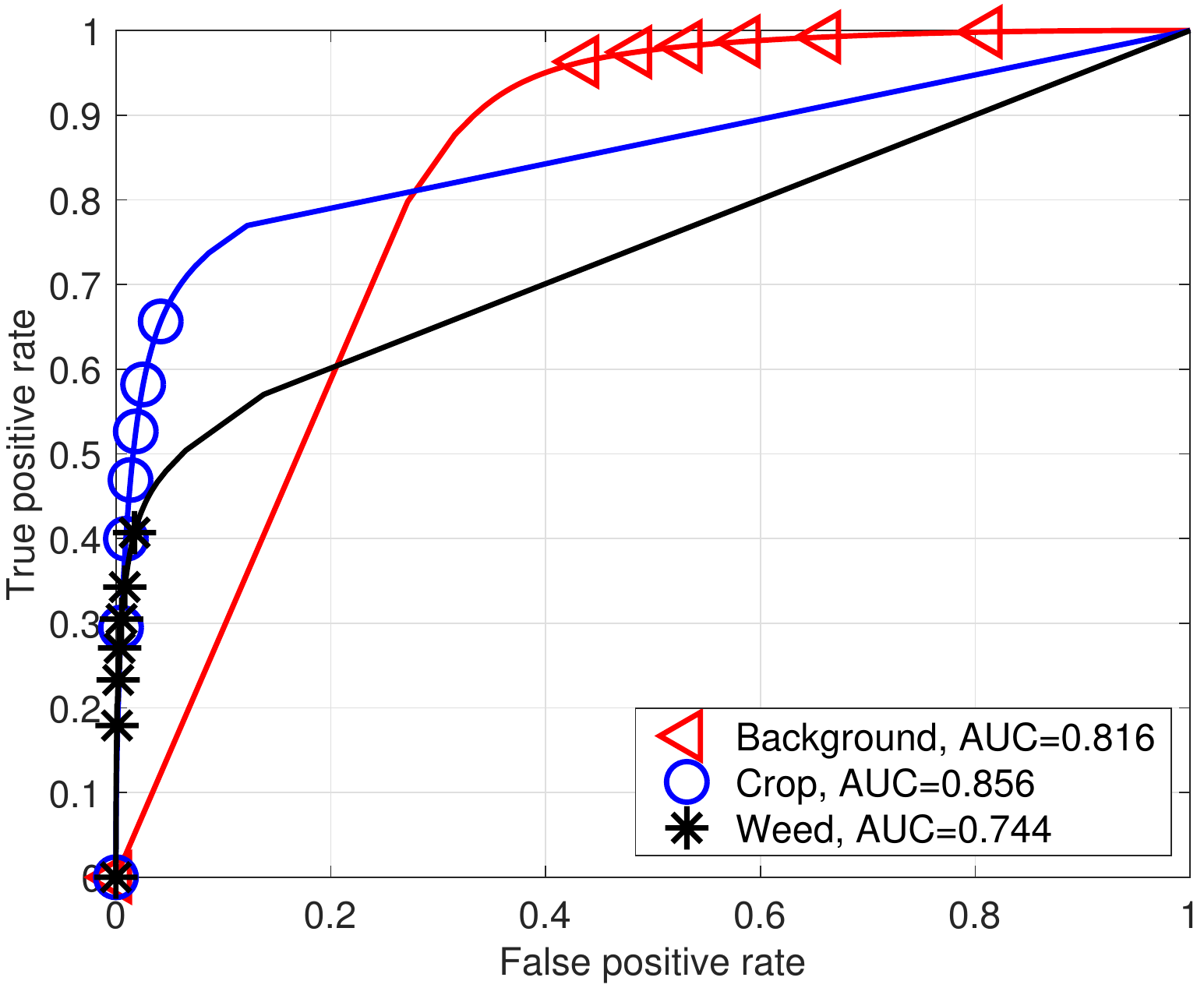}}
\subfloat[]{\includegraphics[width=\picWidth\columnwidth]{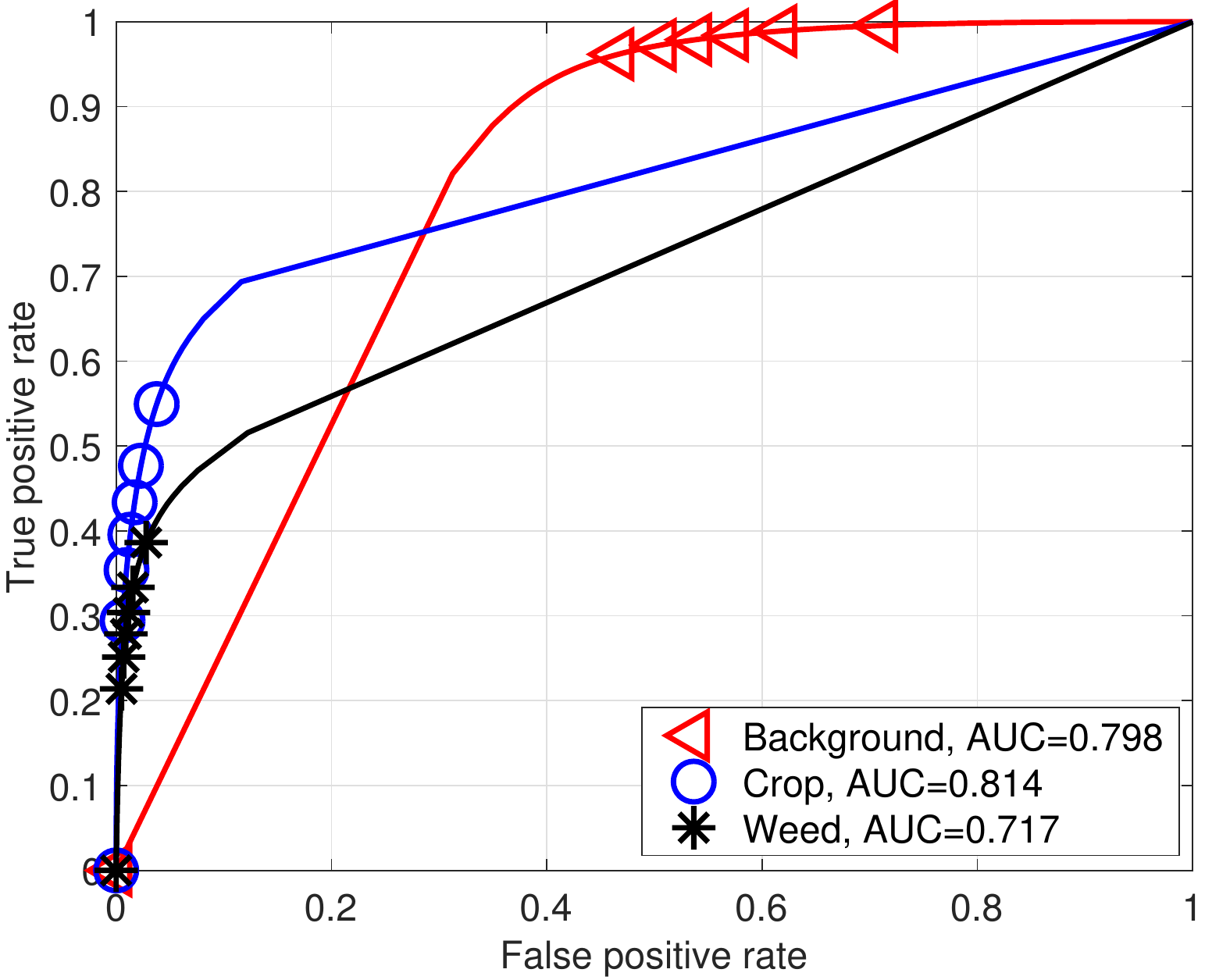}}
\subfloat[]{\includegraphics[width=\picWidth\columnwidth]{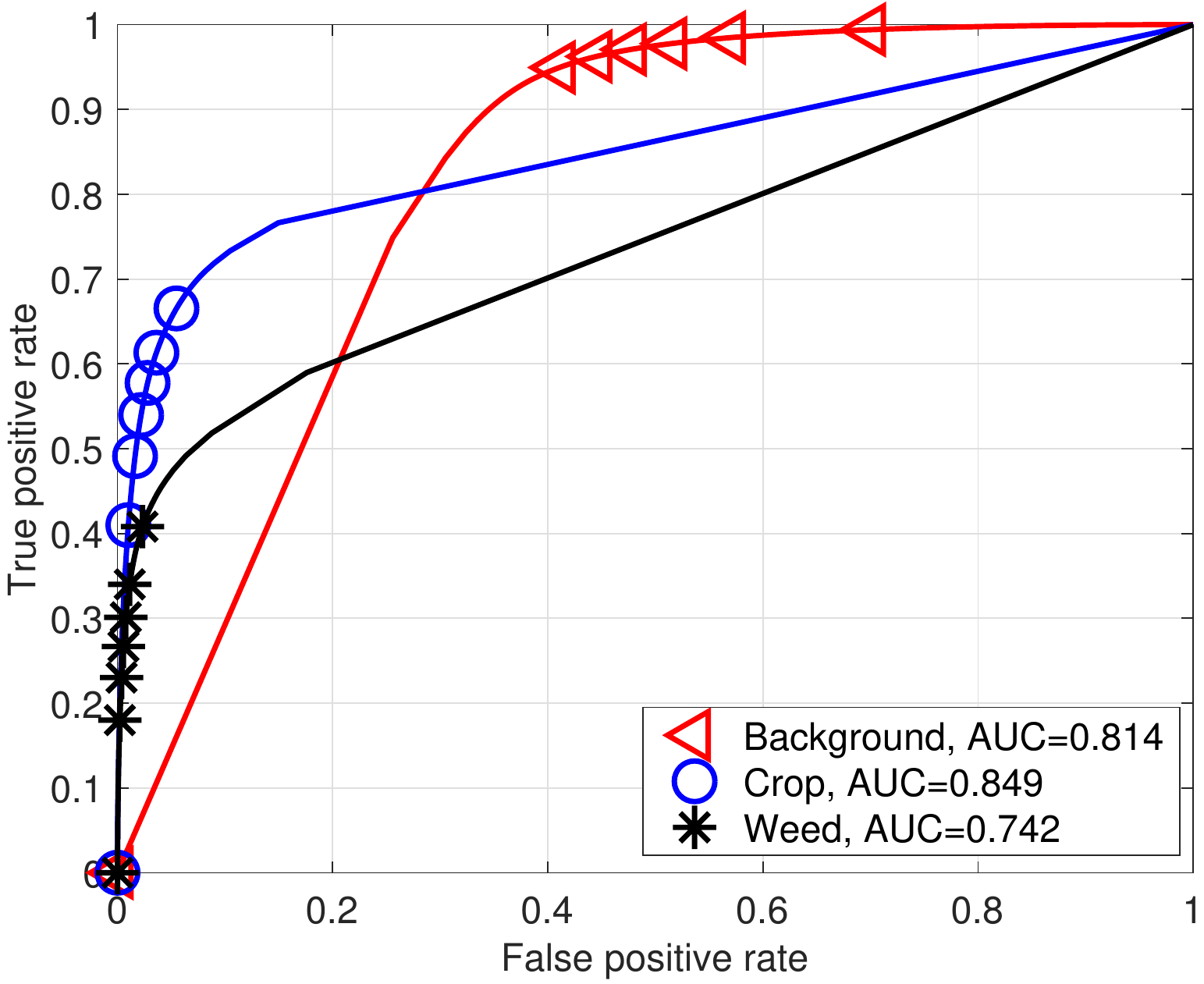}}
\subfloat[]{\includegraphics[width=\picWidth\columnwidth]{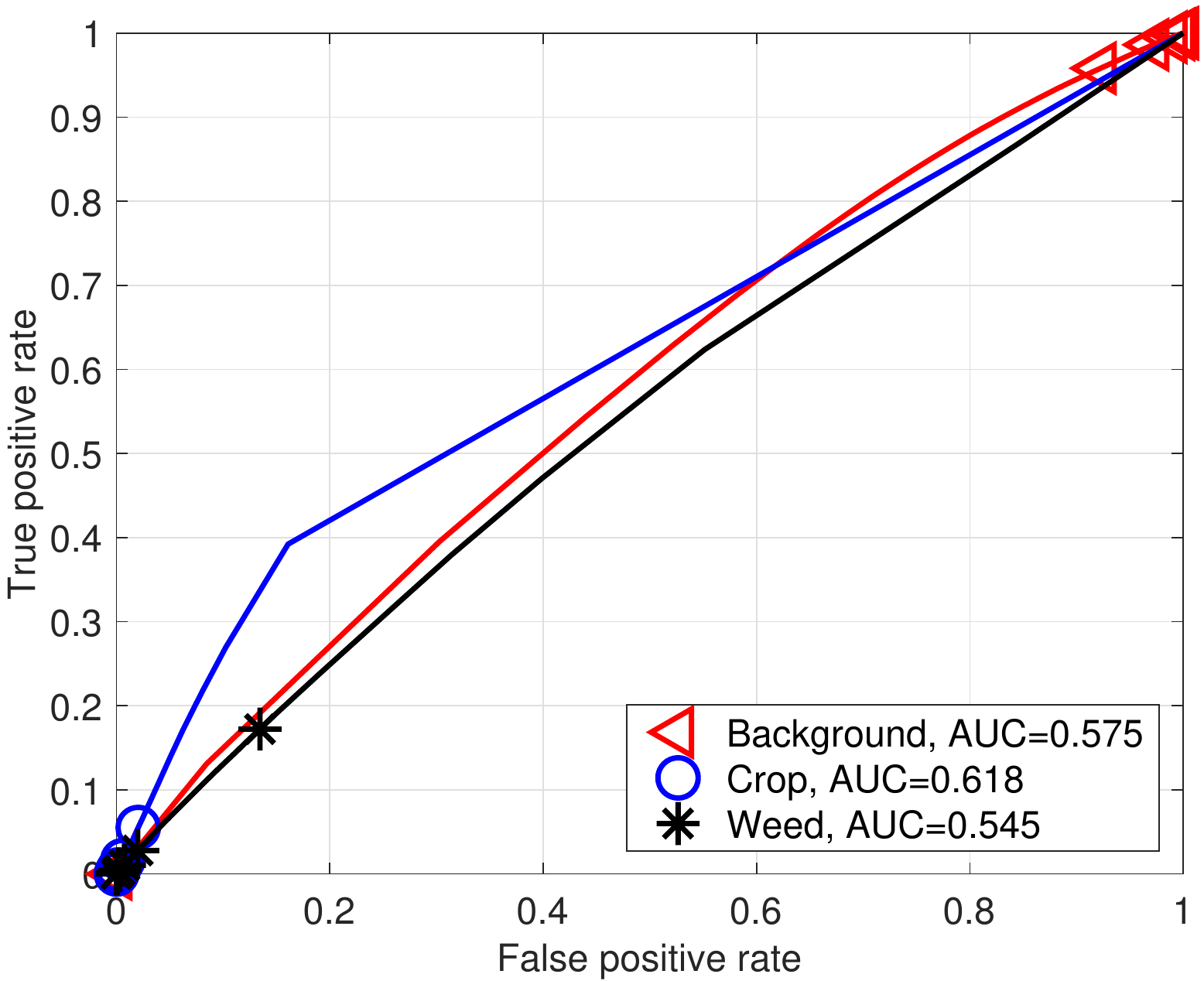}}

\subfloat[]{\includegraphics[width=\picWidth\columnwidth]{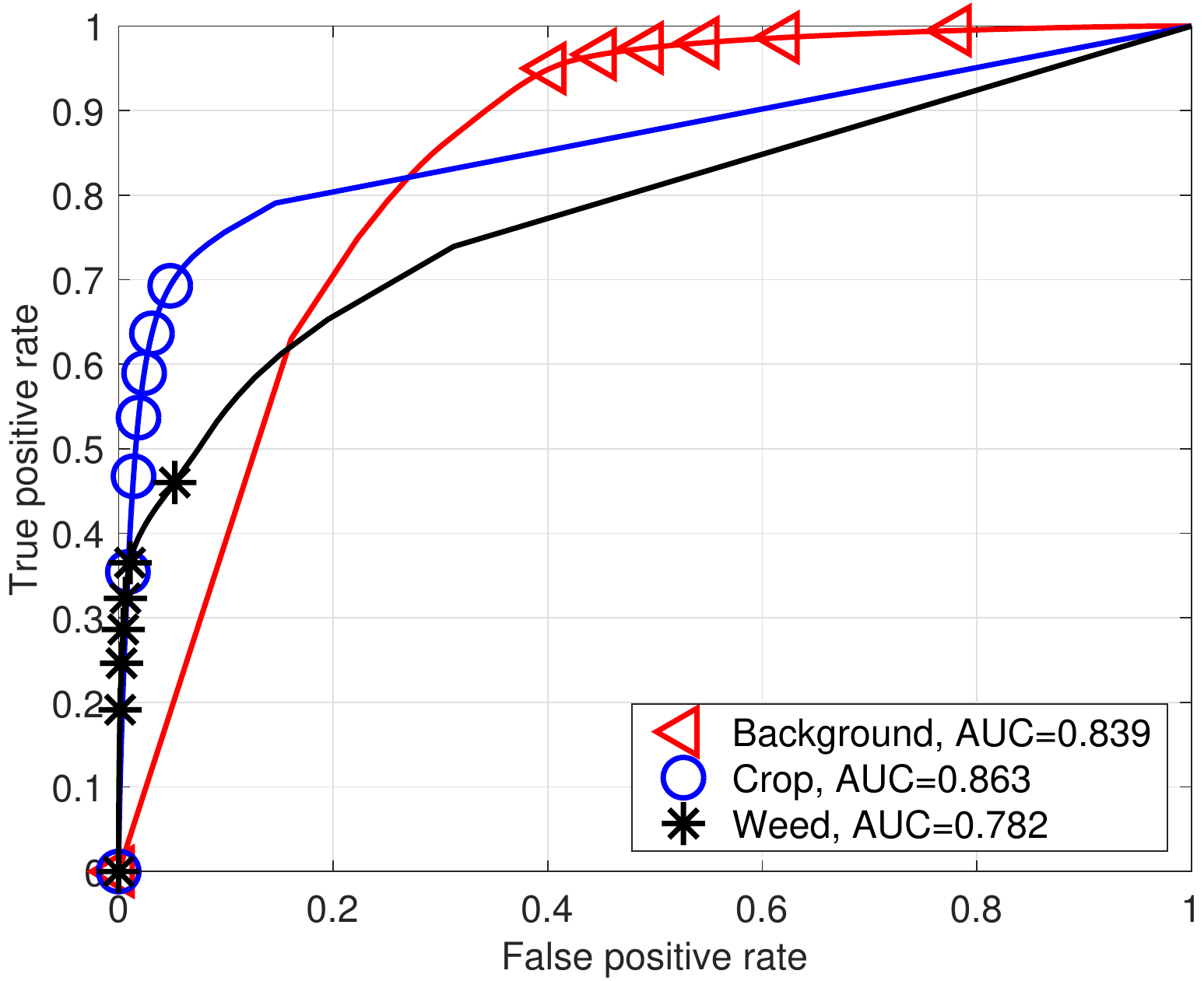}}
\subfloat[]{\includegraphics[width=\picWidth\columnwidth]{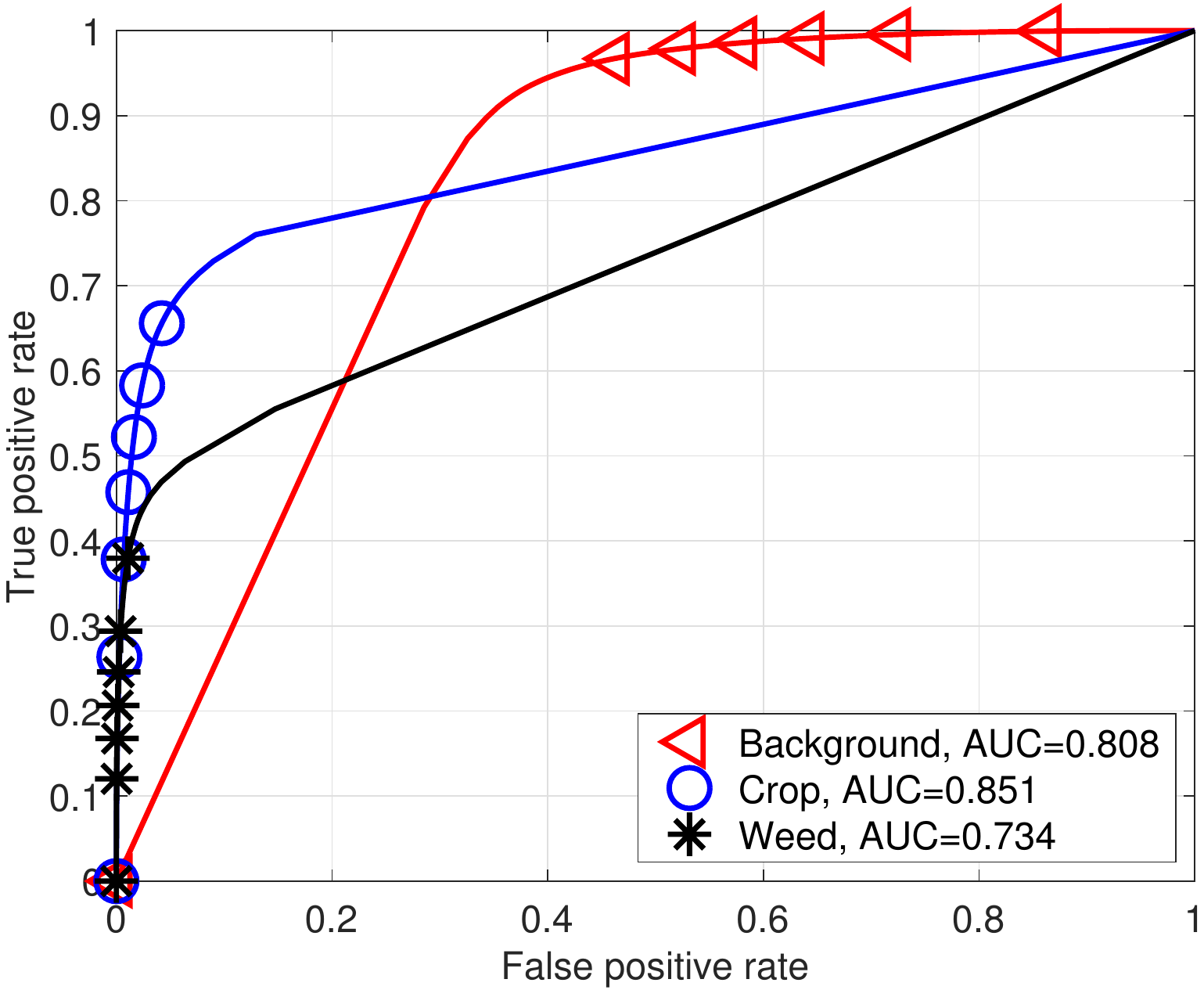}}
\subfloat[]{\includegraphics[width=\picWidth\columnwidth]{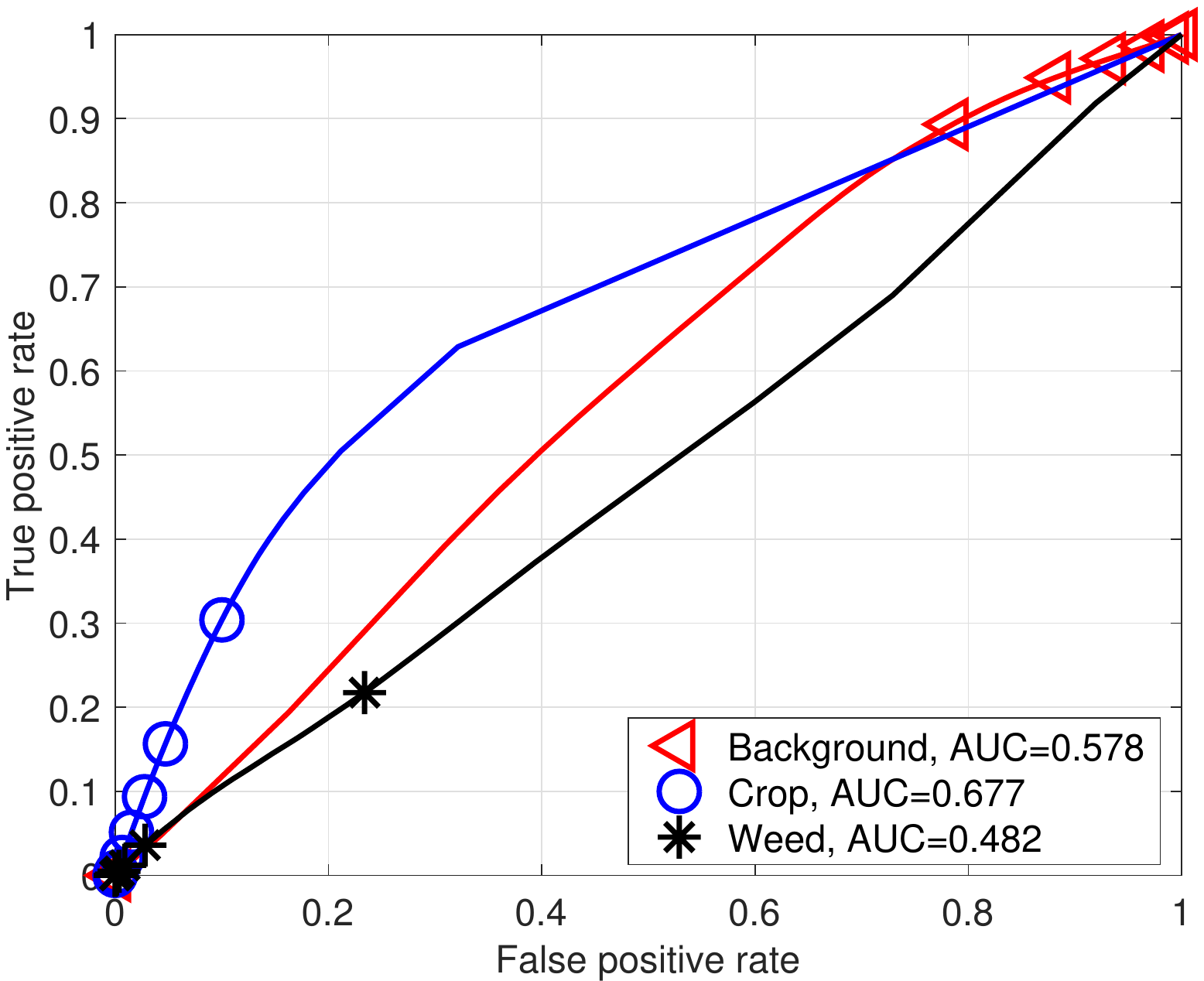}}
\subfloat[]{\includegraphics[width=\picWidth\columnwidth]{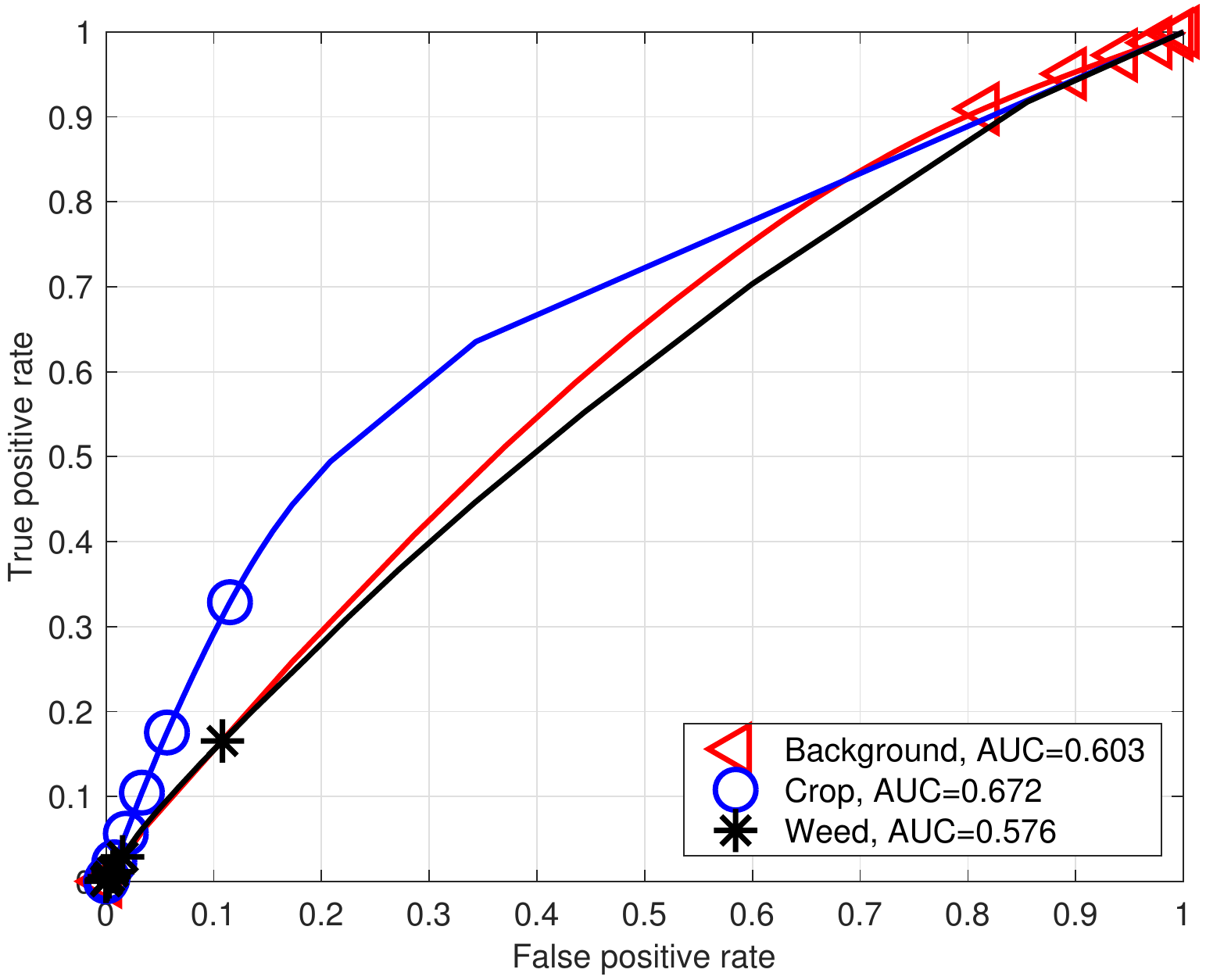}}

\subfloat[]{\includegraphics[width=\picWidth\columnwidth]{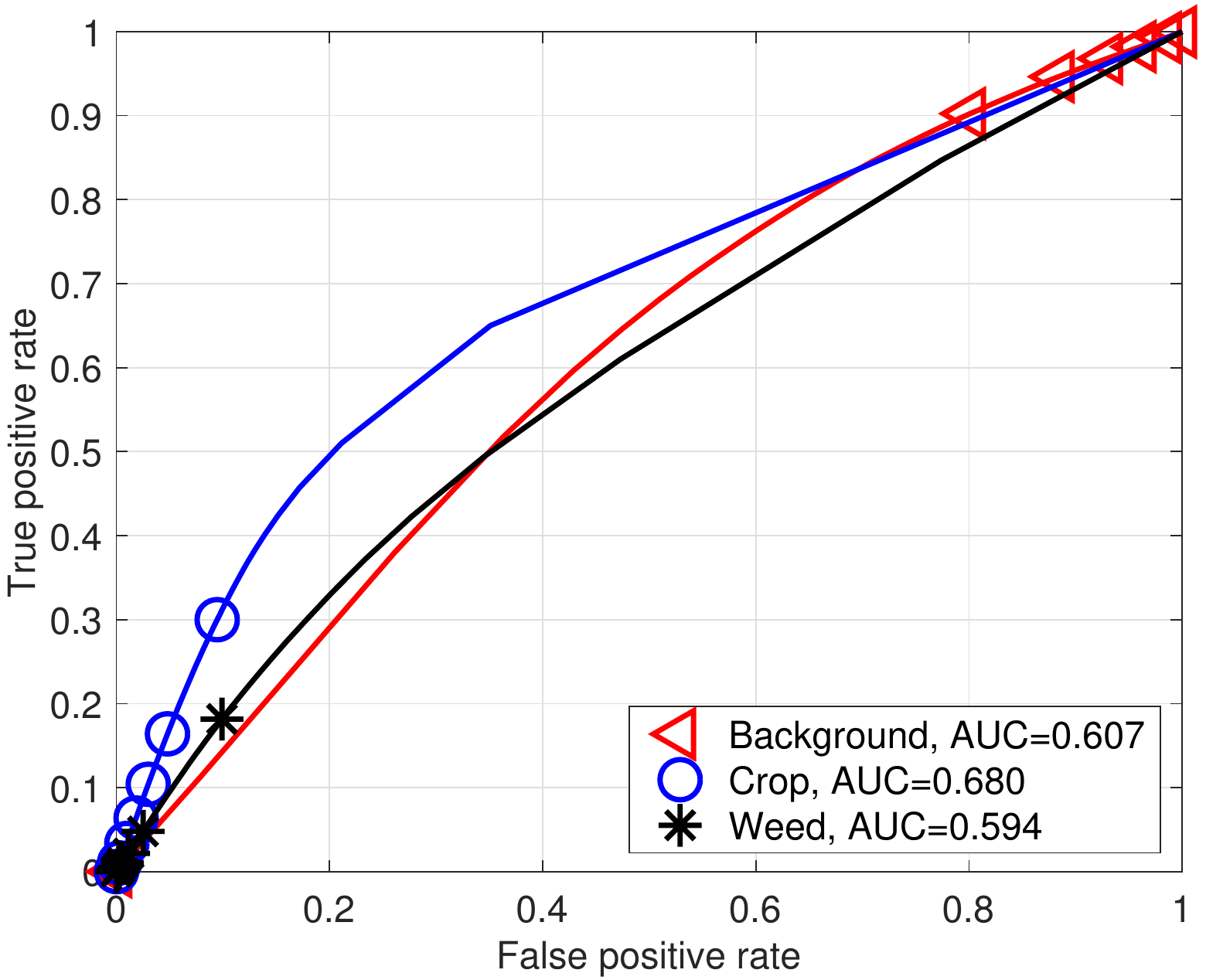}}
\subfloat[]{\includegraphics[width=\picWidth\columnwidth]{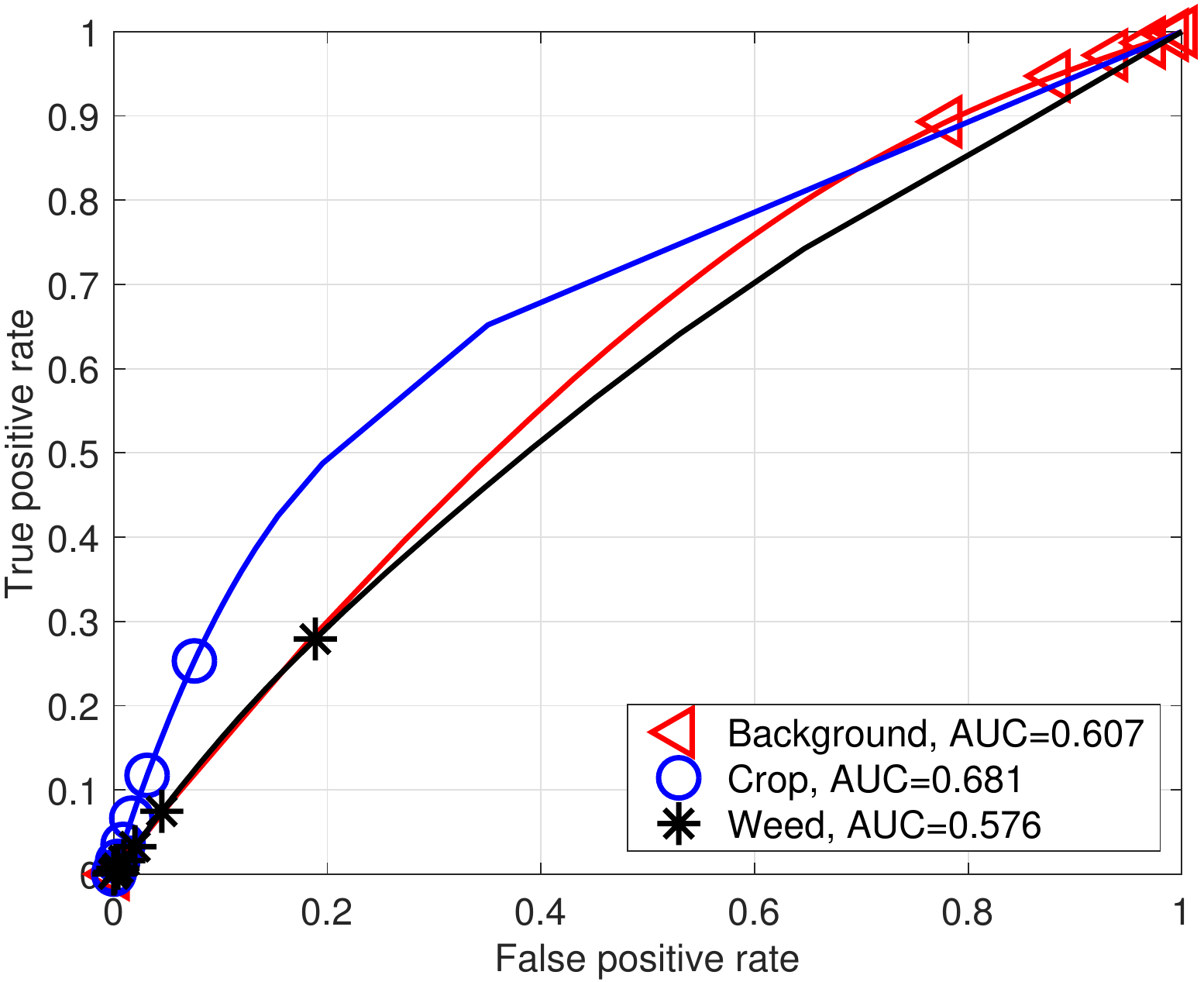}}
\subfloat[2]{\includegraphics[width=\picWidth\columnwidth]{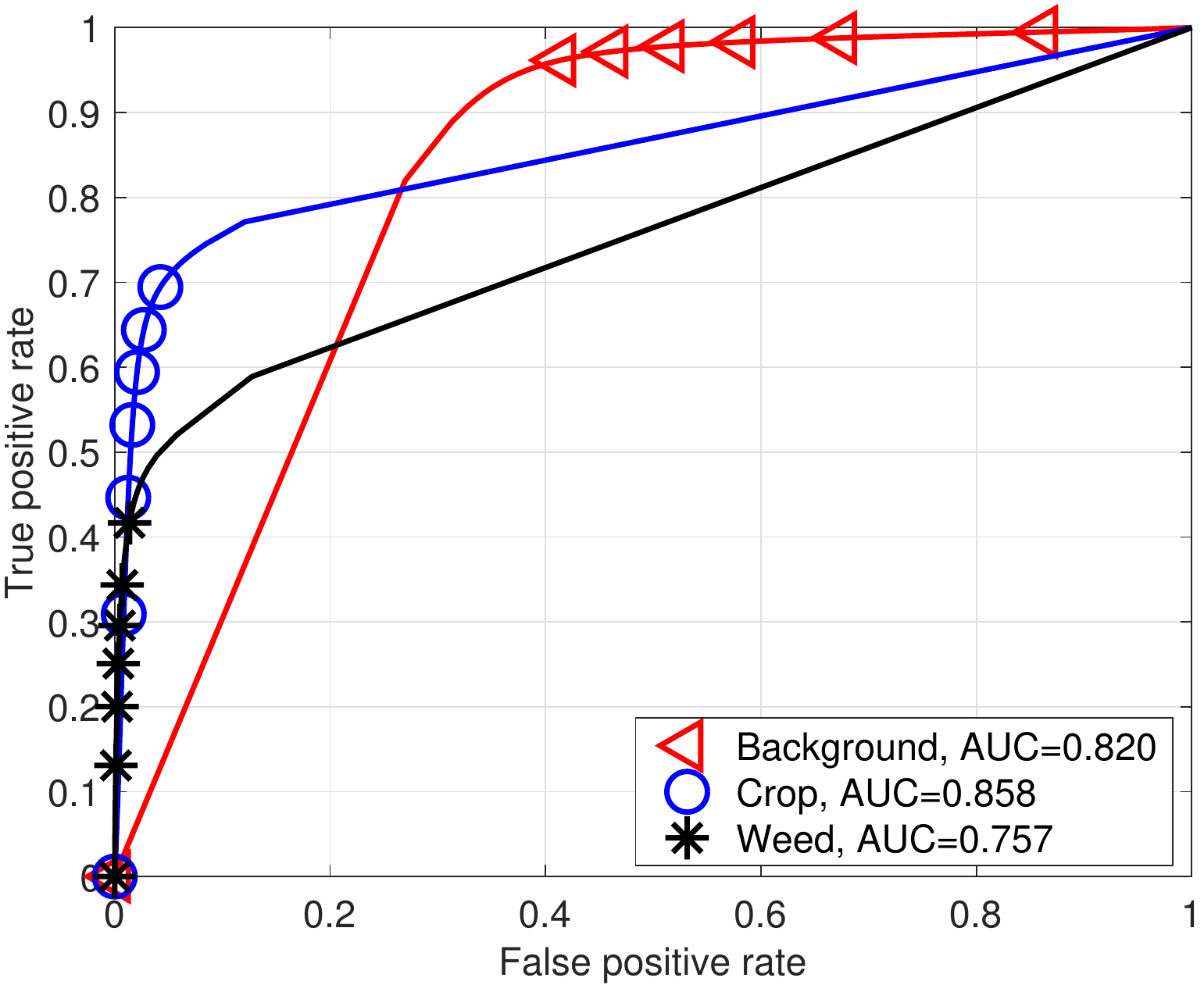}}
\subfloat[]{\includegraphics[width=\picWidth\columnwidth]{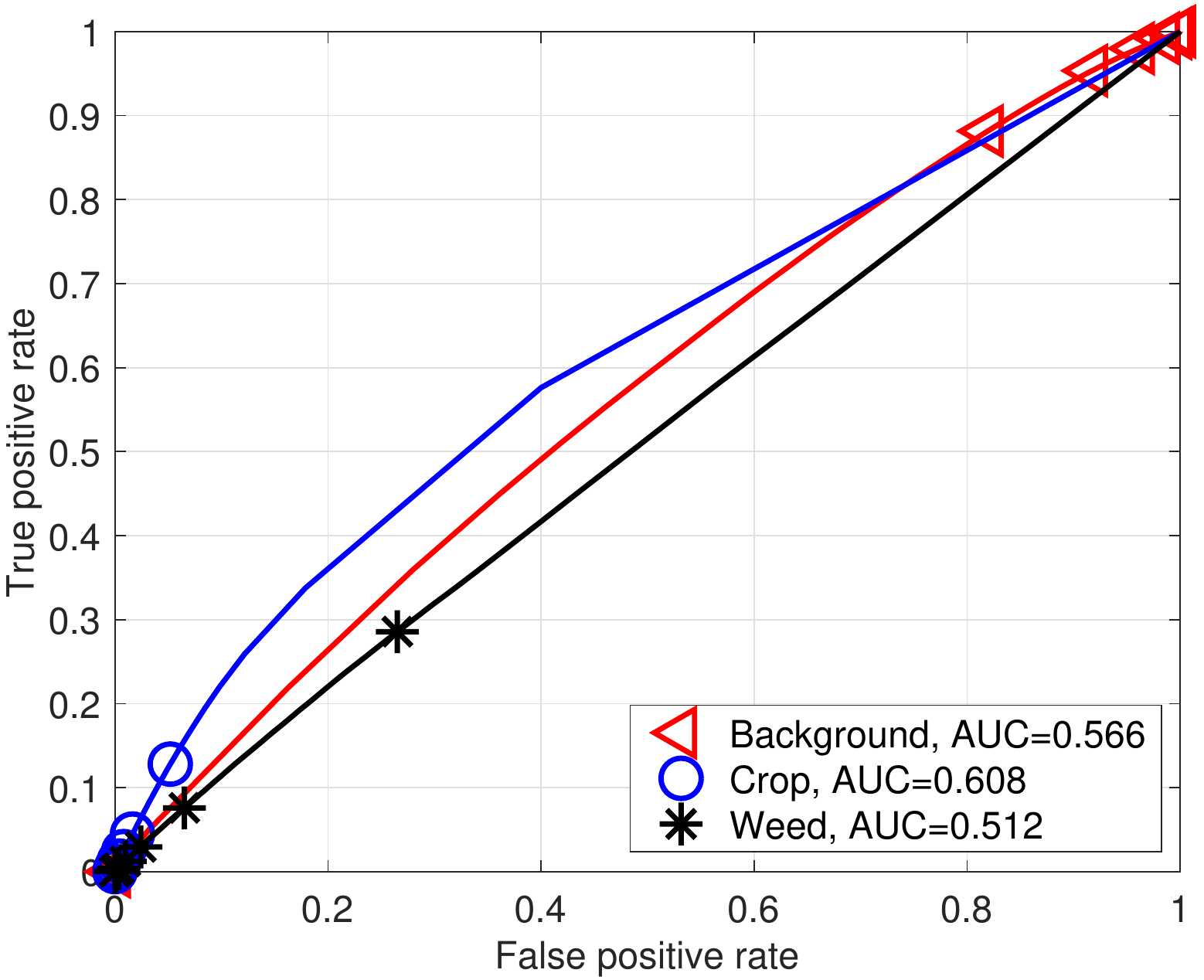}}
    \caption{Performance curves of the RedEdge-M dataset models (Model 1--13). 
    For improved visualization, note that we intentionally omit Model 11, which performs very similarly to Model~10. (\textbf{a})~Model~1; (\textbf{b}) Model 2; (\textbf{c}) Model 3; (\textbf{d}) Model 4; (\textbf{e}) Model 5; (\textbf{f}) Model 6; (\textbf{g}) Model 7; (\textbf{h}) Model 8; (\textbf{i})~Model 9; (\textbf{j}) Model 10; (\textbf{k}) Model 12; (\textbf{l}) Model 13.}
    \label{fig:rededge-AUC-curves}
\end{figure}  \unskip

\begin{figure}[H]
\centering
\includegraphics[width=\columnwidth]{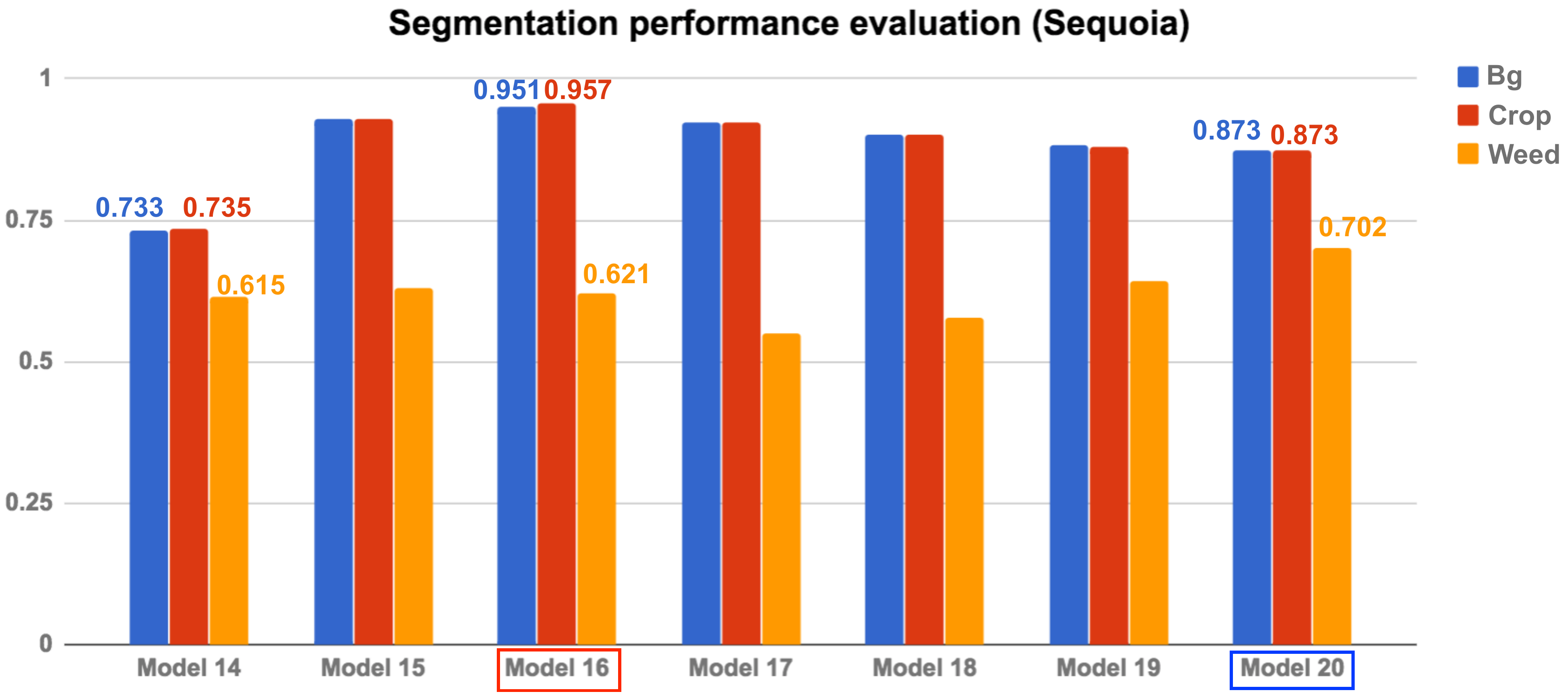}
    \caption{Quantitative evaluation of the segmentation using area under the curve (AUC) of the Sequoia dataset. As in the RedEdge-M dataset, the red box indicates the best model, and the blue box~is a model with only one NDVI image input.}
    \label{fig:sequoia-AUC-bars}
\end{figure}

\def\picWidth{0.25}
\begin{figure}[H]
\centering
\subfloat[]{\includegraphics[width=\picWidth\columnwidth]{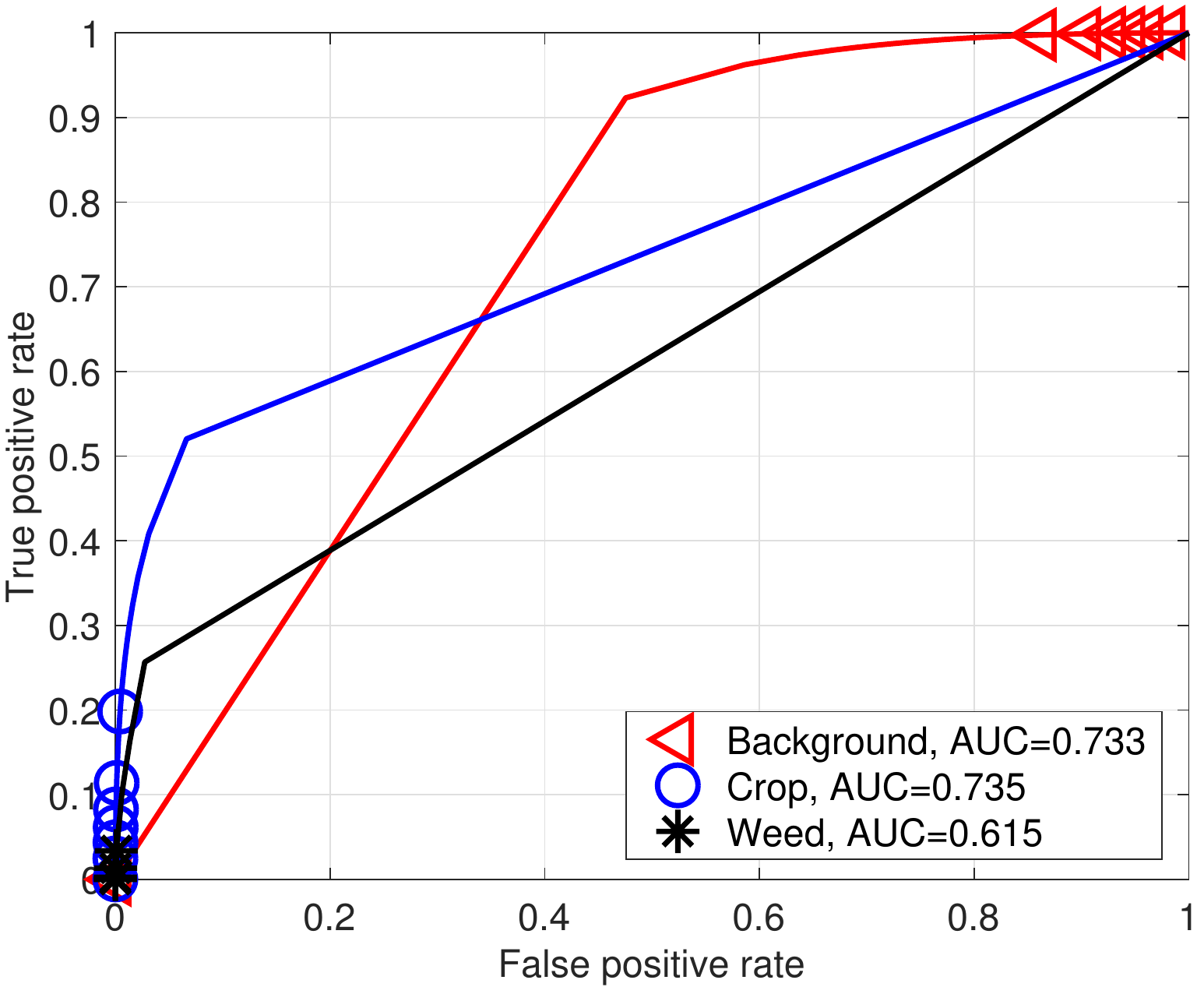}}
\subfloat[]{\includegraphics[width=\picWidth\columnwidth]{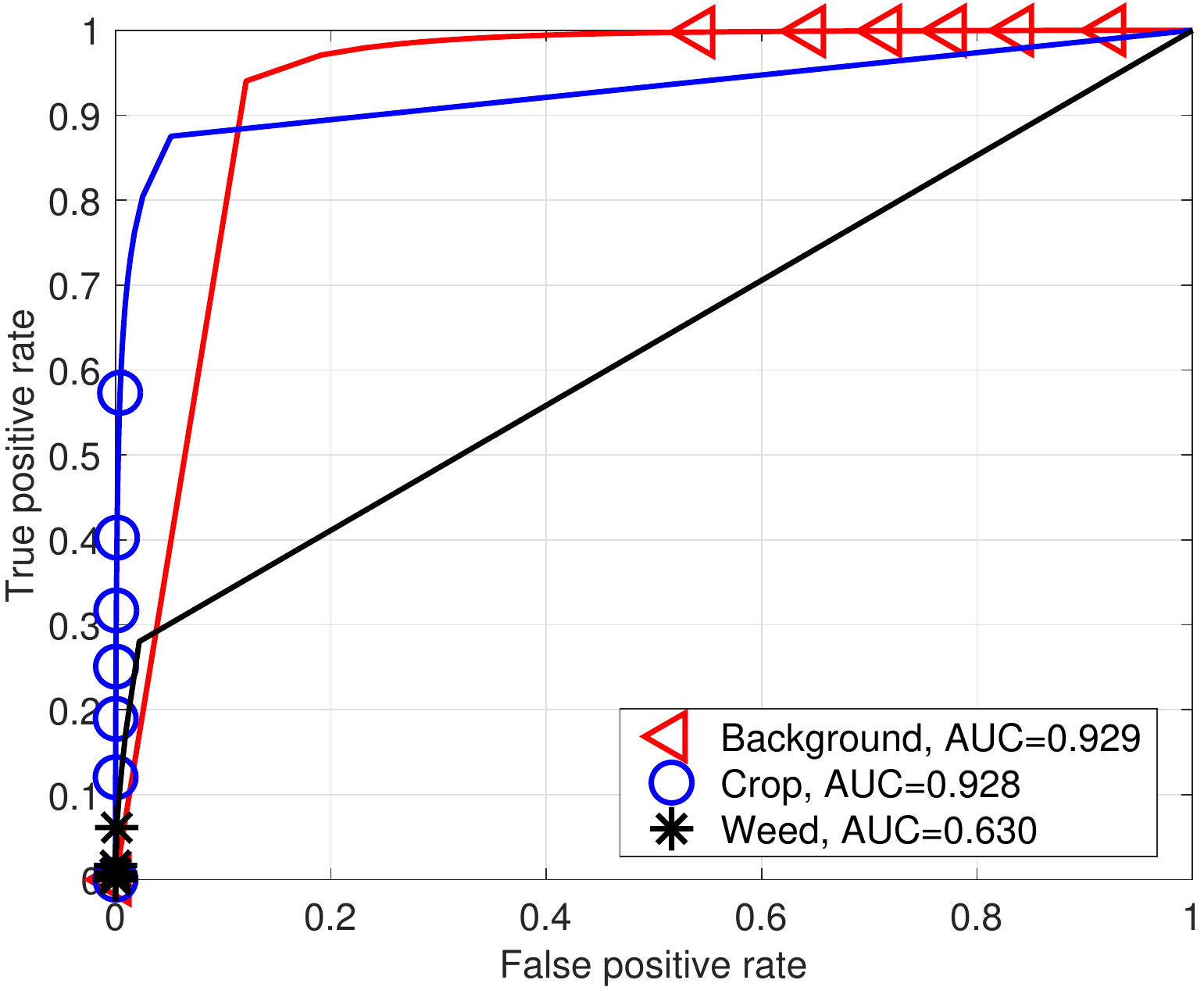}}
\subfloat[Model 16]{\includegraphics[width=\picWidth\columnwidth]{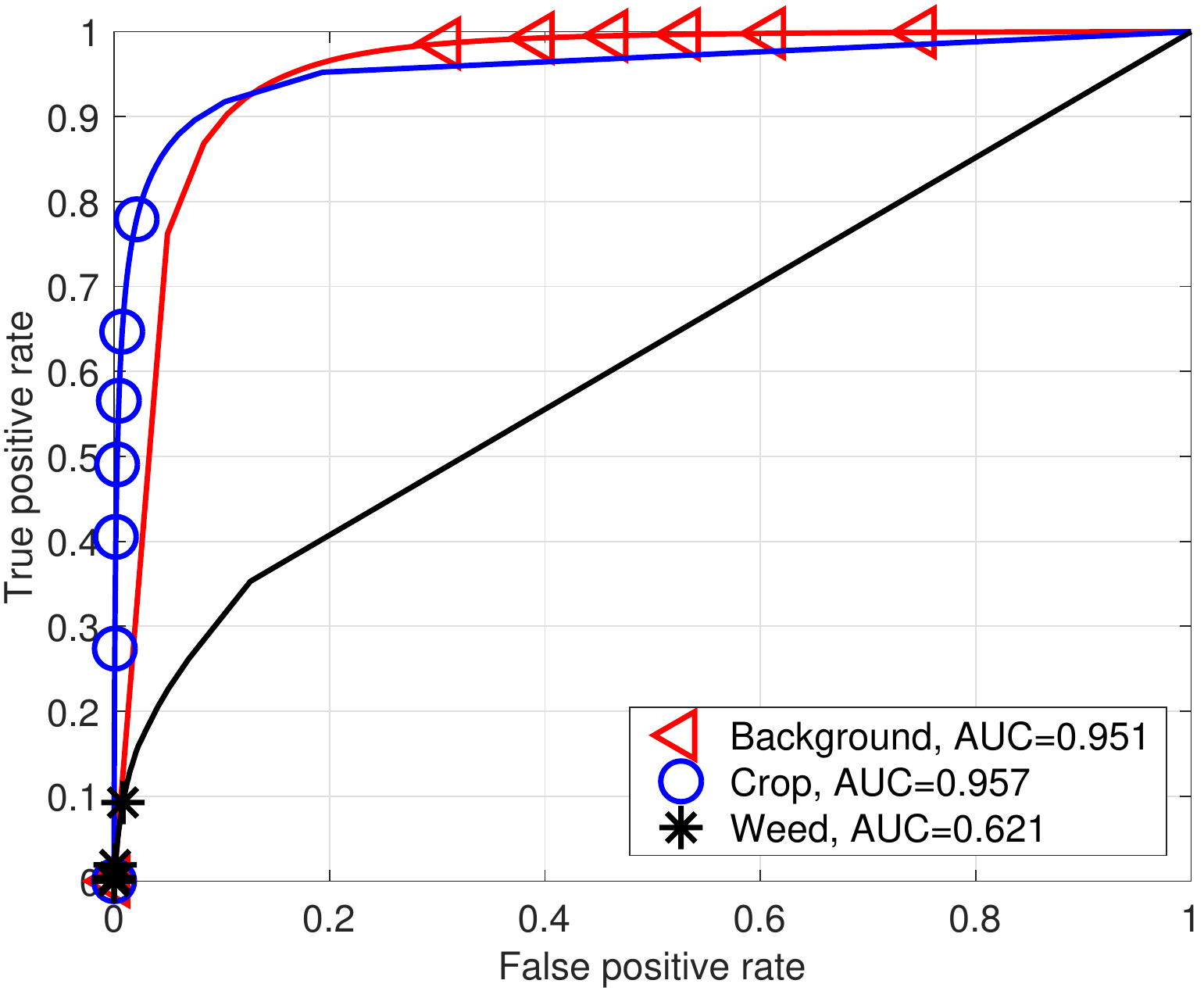}}
\subfloat[Model 17]{\includegraphics[width=\picWidth\columnwidth]{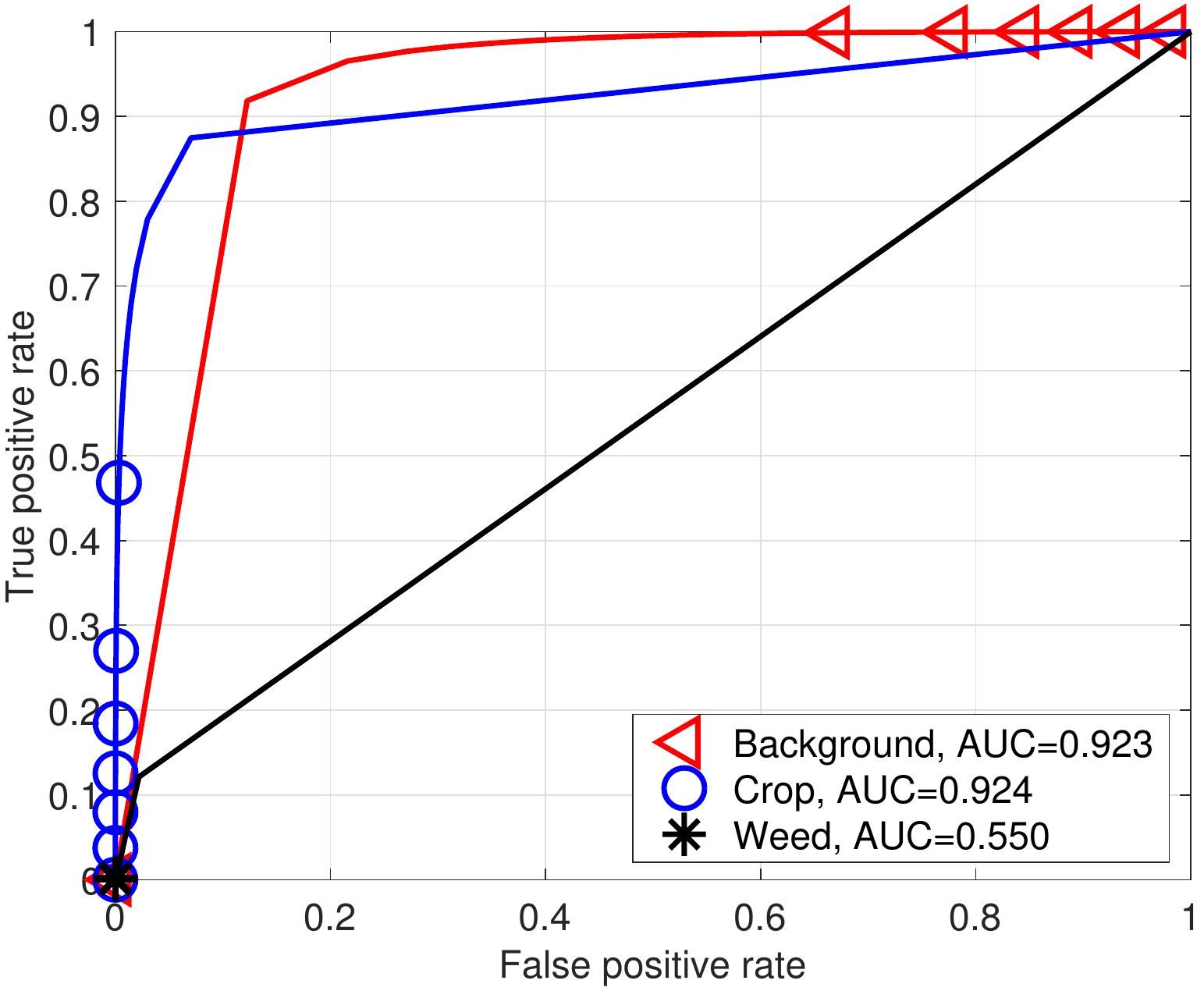}}

\subfloat[Model 18]{\includegraphics[width=\picWidth\columnwidth]{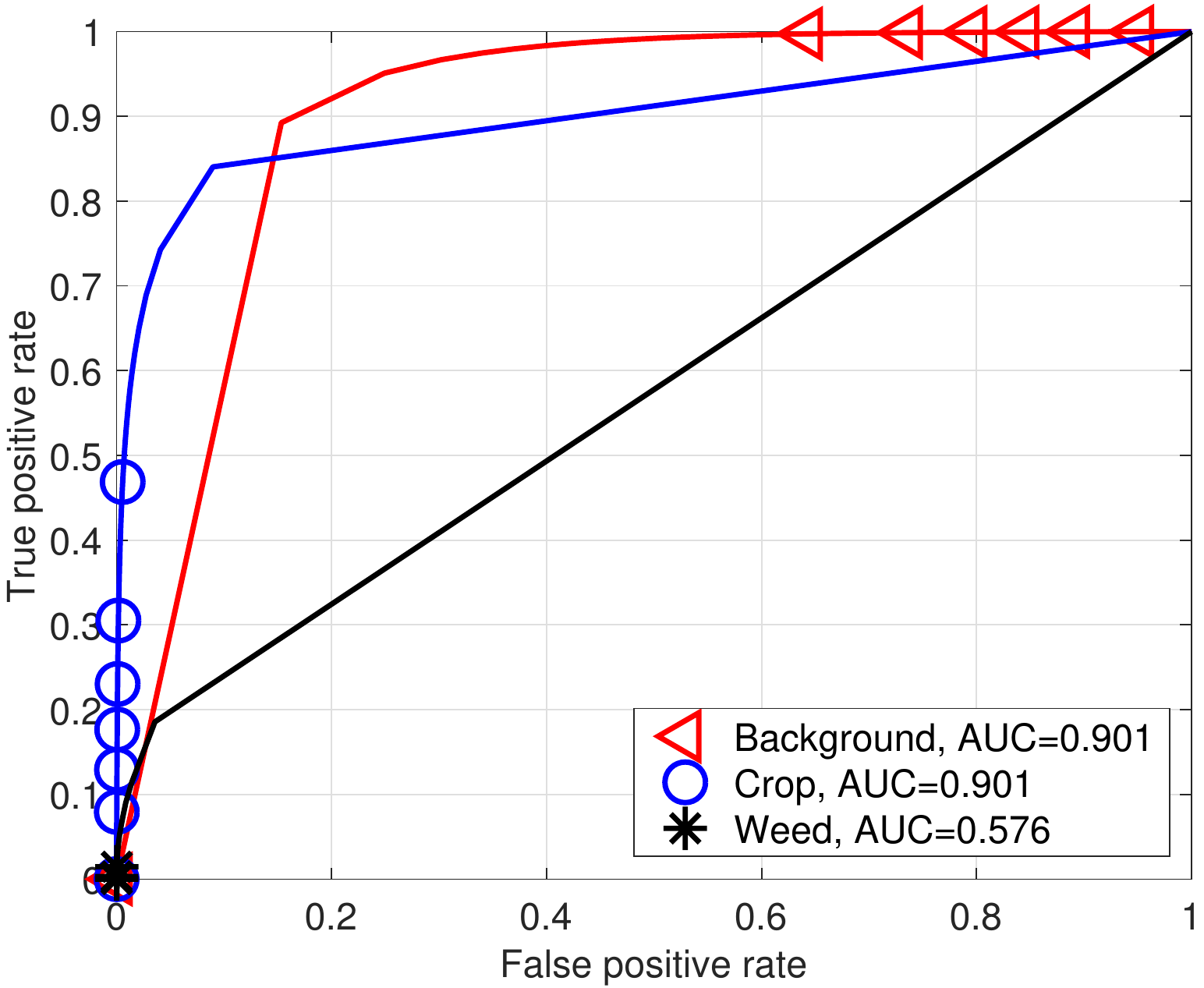}}
\subfloat[Model 19]{\includegraphics[width=\picWidth\columnwidth]{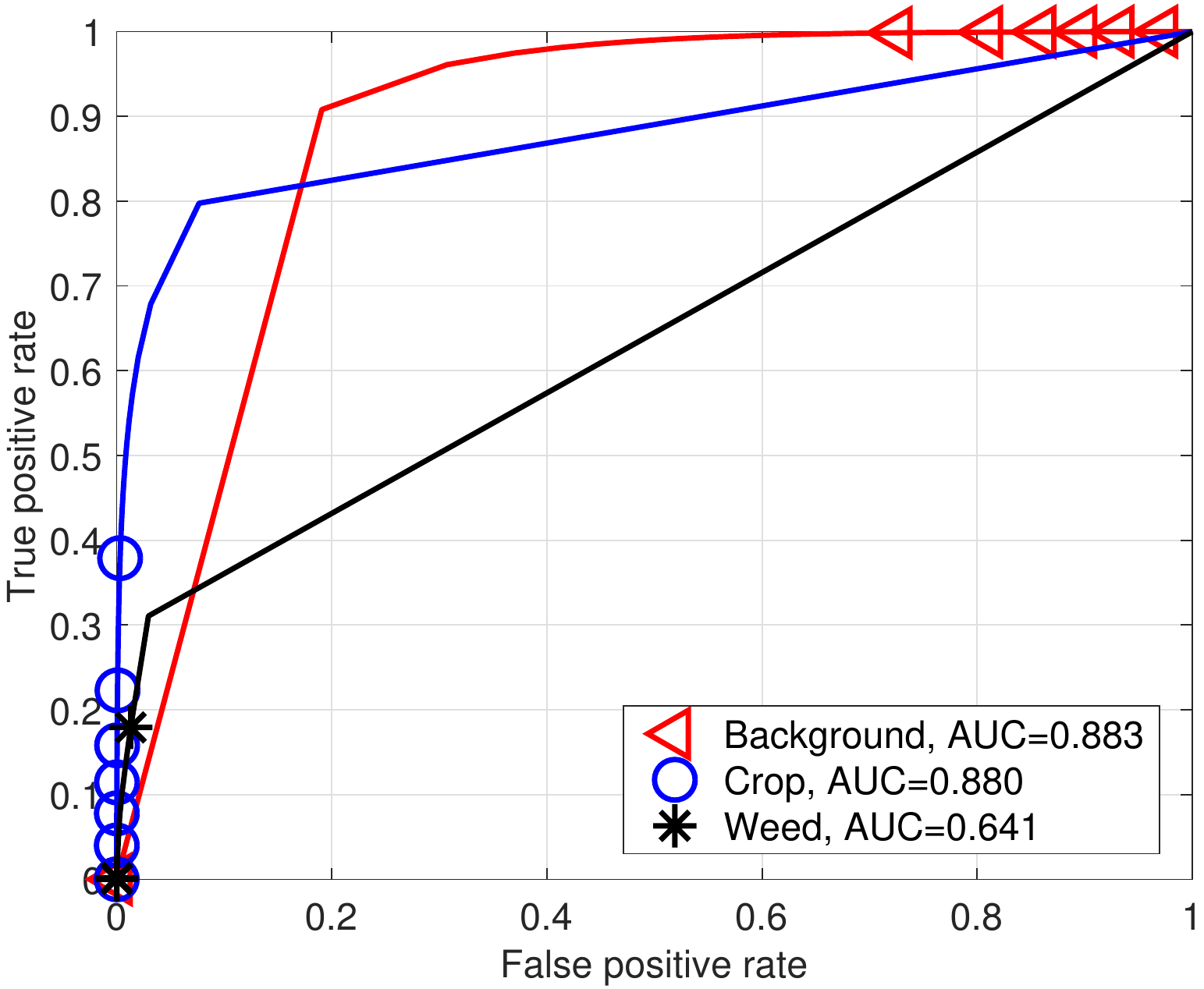}}
\subfloat[Model 20]{\includegraphics[width=\picWidth\columnwidth]{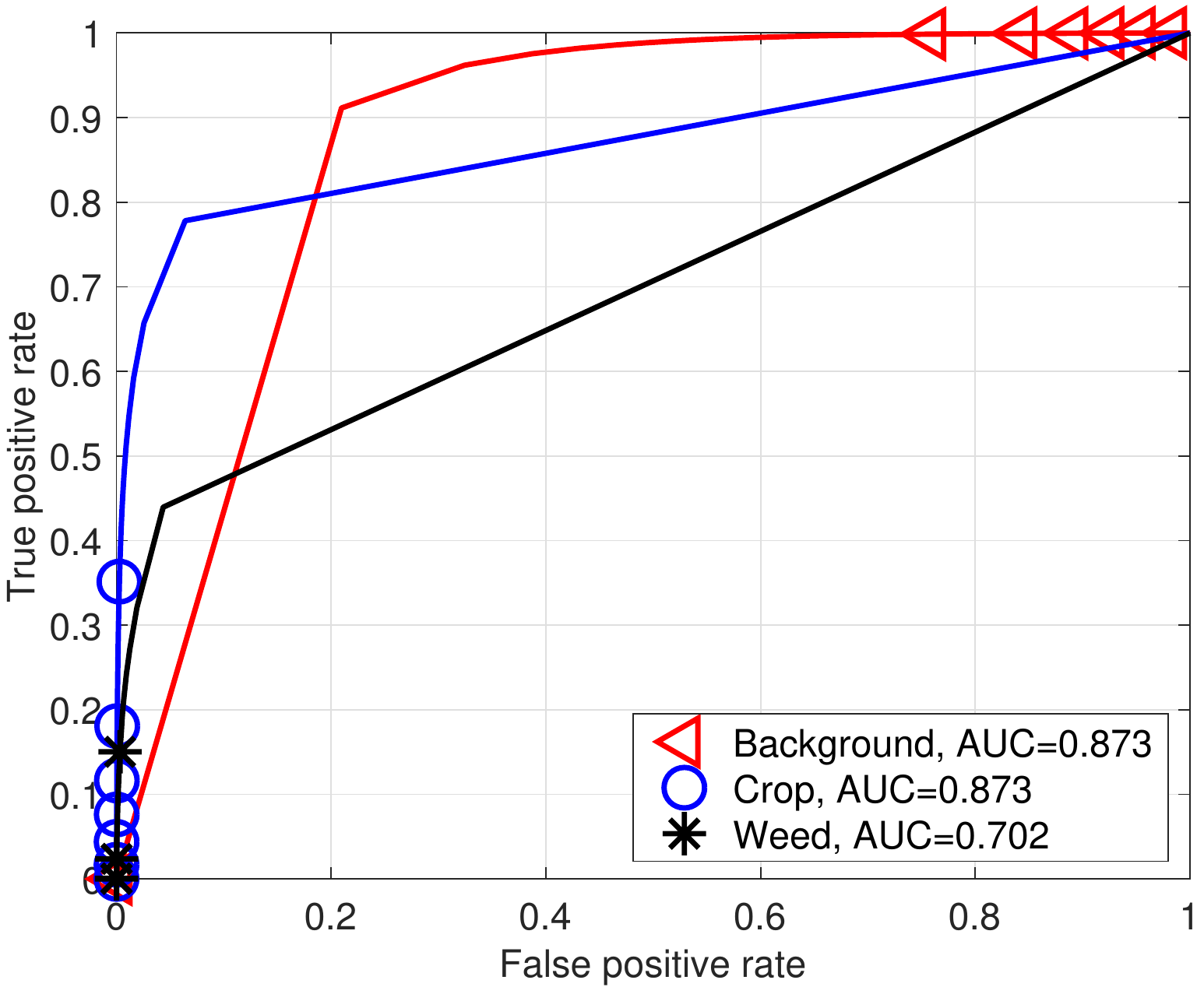}}
    \caption{Performance curves of Sequoia dataset models (Models 14--20). (\textbf{a}) Model 14; (\textbf{b}) Model 15; (\textbf{c}) Model 16; (\textbf{d}) Model 17; (\textbf{e}) Model 18; (\textbf{f}) Model 19; (\textbf{g}) Model 20.}
    \label{fig:sequoia-AUC-curves}
\end{figure}

Model 1 and Model 2 show the impact of batch size; clearly, increasing the batch size yields better results. The maximum batch size is determined by the memory capacity of the GPU, being six in our case with NVIDIA Titan X (Santa Clara, CA, USA). However, as we often fail to allocate six batches into our GPU memory, we~use~five~batches~for most model training procedures apart from the batch size comparison tests.

Model 1 and Model 3 demonstrate the impact of class balancing, as mentioned in Section \ref{sec:network}.

Model 4 and Model 12 are particularly interesting, because the former excludes only the NDVI channel from the 12 available channels, while the latter uses only this channel. The performance of the two classifiers is very different; Model 12 substantially outperforms Model 4. This happens because the NDVI band already identifies vegetation so that the classifier must only distinguish between crops and weeds. Since the NDVI embodies a linearity between the NIR and R channels, we expect a CNN with NIR and R input channels to perform similarly to Model 12 by learning this relationship. Our results show that the NDVI contributes greatly towards accurate vegetation classification. This suggests that, for a general segmentation task, it is crucial to exploit input information which effectively captures distinguishable features between the target classes.

Figures~\ref{fig:rededge-AUC-bars} and \ref{fig:rededge-AUC-curves} show the AUC scores for each model and their performance curves. Note that there are sharp points (e.g., around the 0.15 false positive rate for \texttt{crop} in Figure~\ref{fig:rededge-AUC-curves}c,d). These~are points where neither $\text{precision}_{\mbox{\tiny c}}$ nor $\text{recall}_{\mbox{\tiny c}}$ are changed even with varying thresholds, ($0\leq\epsilon\leq1$). In~this~case, \texttt{perfcurve} generates the point at (1,1) for a monotonic function, enabling AUC to be computed. This~rule~is equally applied to all other evaluations for a fair comparison, and it is obvious that a better classifier should generate a higher precision and recall point, which, in turn, yields~higher AUC even with the linear monotonic function.

\subsubsection{Quantitative Results for the Sequoia Dataset}
For the Sequoia dataset, we train seven models with varying conditions. Our results confirm the trends discussed for the RedEdge-M camera in the previous section. Namely, the NDVI plays a significant role in crop/weed detection. 

Compared to the RedEdge-M dataset, the most noticeable difference is that the performance gap between crops and weeds is more significant due to their small sizes. As described in Section~\ref{sec:data-collection}, the~crop~and weed instances in the RedEdge-M dataset are about 15--20 pixels and 5--10 pixels, respectively. In the Sequoia dataset, they are smaller, as the data collection campaign was carried out at an earlier stage of crop growth. This also reduces weed detection performance (10\% worse weed detection), as expected.

In addition, the Sequoia dataset contains 2.6 times less weeds, as per the class weighting ratio described in Section \ref{sec:data-collection} (the~RedEdge-M dataset has $w_{\mbox{\tiny c}}=[0.0638, 1.0, 1.6817]$ for the [\texttt{bg}, \texttt{crop}, \texttt{weed}] classes, while the Sequoia dataset has [0.0273, 1.0, 4.3802]). This is evident by comparing Model 14 and Model 15, as the former significantly outperforms without class balancing. These results are contradictory to those obtained from Model 1 and Model 3 from the RedEdge-M dataset, as Model 1 (with class balancing) slightly outperforms Model 3 (without class balancing). Class balancing can therefore yield both advantages and disadvantages, and its usage should be guided by the datasets and application at hand.

\subsection{Qualitative Results}
Alongside the quantitative performance evaluation, we also present a qualitative analysis in Figures ~\ref{fig:rededgeQuantResults} and \ref{fig:sequoiaQuantResults} for the RedEdge-M and Sequoia testing datasets, i.e., datasets 003 and 005. We use the best performing models (Model 5 for RedEdge-M and Model 16 for Sequoia) {that reported AUC of [0.839, 0.863, 0.782] for RedEdge-M and [0.951, 0.957, 0.621] for Sequoia in order }to generate the results. As high-resolution images are hard to visualize {due to technical limitations such as display or printer resolutions}, we display center aligned images with varying zoom levels (17\%, 25\%, 40\%, 110\%, 300\%, 500\%) in Figure ~\ref{fig:rededgeQuantResults}a--f. The columns correspond to input images, ground truth images, and the classifier predictions. The color convention follows \color{blue} bg, \color{CommentDG} crop, \color{red} weed \color{black}. 

\begin{figure}[H]
\centering
\includegraphics[width=\columnwidth]{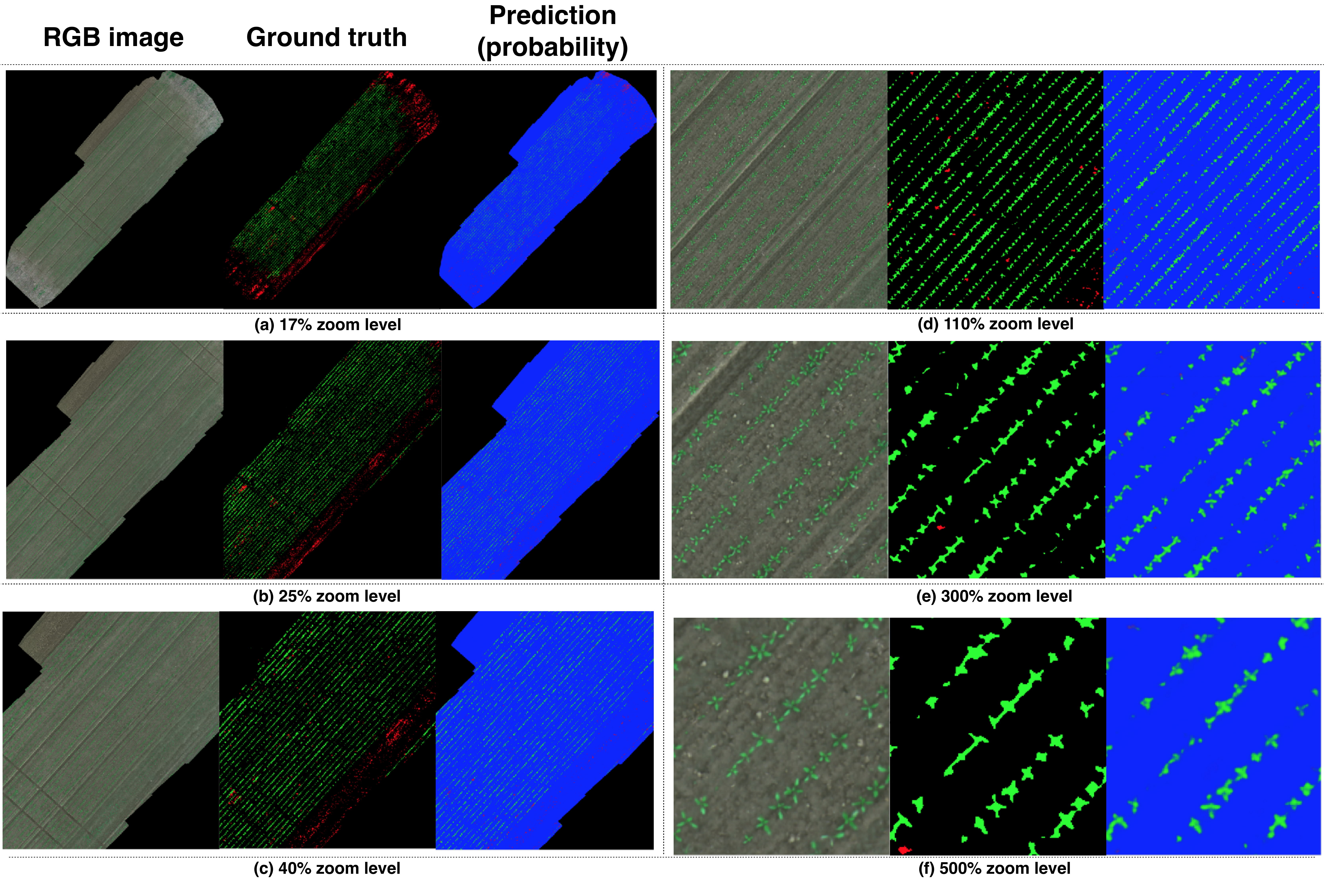}
    \caption{Quantitative results for the RedEdge-M testing dataset (dataset 003). Each column corresponds to an example input image, ground truth, and the output prediction. Each row (\textbf{a}--\textbf{f}) shows a different zoom level on the orthomosaic weed map. The color convention follows bg, crop, weed. These images are best viewed in color.}
    \label{fig:rededgeQuantResults}
\end{figure}

\begin{figure}[H]
\centering
\includegraphics[width=\columnwidth]{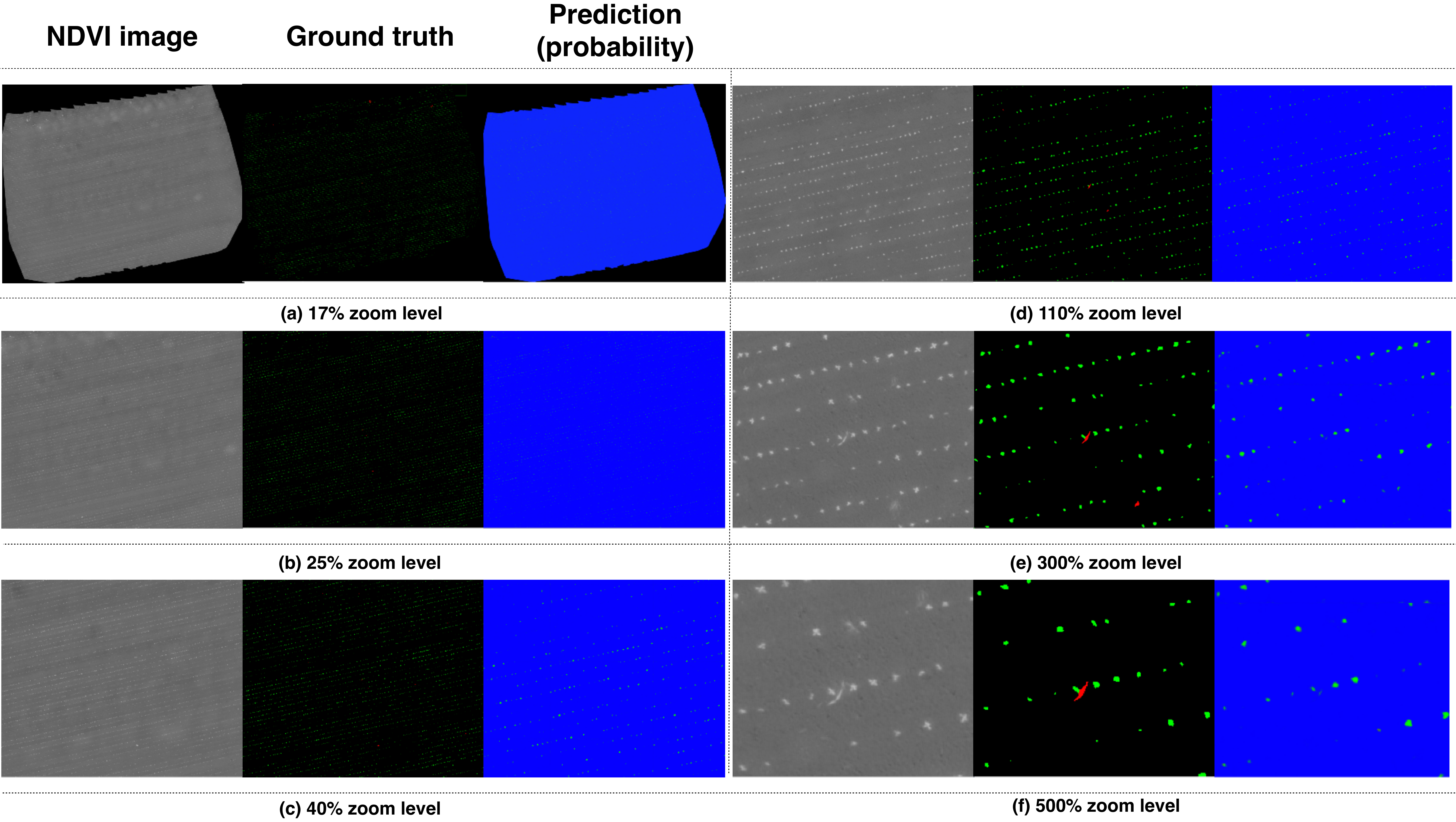}
    \caption{Quantitative results for the Sequoia testing dataset (dataset 005). Each column corresponds to an example input image, ground truth, and the output prediction. Each row (\textbf{a}--\textbf{f}) shows a different zoom level on the orthomosaic weed map. The color convention follows \color{blue} bg, \color{CommentDG} crop, \color{red} weed \color{black}. These images are best viewed in color.}
    \label{fig:sequoiaQuantResults}
        \vspace{-3mm}
\end{figure}


\subsubsection{RedEdge-M Analysis}
In accordance with the quantitative analysis in Table~\ref{tbl:resultSummary}, the classifier performs reasonably for crop prediction. Figure~\ref{fig:rededgeQuantResults}b,c show crop rows clearly, while their magnified views in Figure \ref{fig:rededgeQuantResults}e,f reveal visually accurate performance at a higher resolution.

Weed classification, however, shows {relatively} inferior performance in terms of false positives (wrong detections) and false negatives (missing detections) {than crop classification}. In wider views (e.g., Figure~\ref{fig:rededgeQuantResults}a--c), it can be seen that the weed distributions and their densities are estimated correctly. The rightmost side, top, and bottom ends of the field are almost entirely occupied by weeds, and prediction reports consistent results with high precision but low recall.

This behavior is likely due to several factors. Most importantly, the weed footprints in the images are too small to distinguish, as exemplified in Figure~\ref{fig:rededgeQuantResults}e,f in the lower left corner. Moreover, the total number of pixels belonging to \texttt{weed} are relatively smaller than those in \texttt{crop}, which  implies limited weed instances in the training dataset. Lastly, the testing images are unseen as they were recorded~in different sugar beet fields than the training examples. This implies that our classifier could be overfitting to the training dataset, or that it learned from insufficient training data for weeds, which only represent a small portion of their characteristics. The higher AUC for crop classification supports this argument, as this class holds less variable attributes across the farm fields. 

\subsubsection{Sequoia Analysis}
The qualitative results for the Sequoia dataset portray similar trends as those of the RedEdge-M dataset, i.e., good crop and relatively poor weed predictions. We make three remarks with respect~to the RedEdge-M dataset. Firstly, the footprints of crops and weeds in an image are smaller since data collection was performed at earlier stages of plant growth. Secondly, there was a gap of two weeks between the training (18 May 2017) and testing (5 May 2017) dataset collection, implying a substantial variation in plant size. Lastly, similar to RedEdge-M, the total amount of pixels belonging~to the \texttt{weed} class is fewer than those in \texttt{crop}.

\section{Discussion on Challenges and Limitations}\label{sec:discussion}
As shown in the preceding section and our earlier work \cite{sa2018ral}, obtaining a reliable spatiotemporal model which can incorporate different plant maturity stages and several farm fields remains a challenge. This is because the visual appearance of plants changes significantly during their growth, with~particular difficulties in distinguishing between crops and weeds in the early season when they appear similar. High-quality and high-resolution spatiotemporal datasets are required to address this issue. However, while obtaining such data may be feasible for crops, weeds are more difficult to capture representatively due to the diverse range of species which can be found in a single field.

More aggressive data augmentation techniques, such as image scaling and random rotations, in~addition~to our horizontal flipping procedure, could improve classification performance, as~mentioned~by \cite{Wang2017-ci,Wong2016-pm}. However, these ideas are only relevant for problems where inter-class variations are large. In our task, applying such methods may be counter-productive as the target classes appear visually similar.

In terms of speed, network forward inferencing takes about 200\unit{ms} per input image on a NVIDIA Titan X GPU processor, while total map generation depends on the number of tiles in an orthomosaic map. For example, the RedEdge-M testing dataset (003) took 18.8\unit{s} (94$\times$0.2), while the Sequoia testing dataset (005) took 42\unit{s}. Although this process is performed with a traditional desktop computer, it~can~be~accelerated through hardware (e.g., using a state-of-the-art mobile computing device (e.g., NVIDIA Jetson Xavier~\cite{NVIDIA:xavier}) or software improvements (e.g.,~other~network~architectures). Note that we omit additional post-processing time from the total map generation, including tile loading and saving the entire weed map, because as these are much faster than the forward prediction step.

The weed map generated can provide useful information for creating prescription maps for the sugar beet field, which can then be transferred to automated machinery, such as fertilizer or herbicide boom sprayers. This procedure allows for minimizing chemical usage and labor cost (i.e.,~environmental~and economical impacts) while maintaining the agricultural productivity of the~farm.
\section{Conclusions}\label{sec:conclusion}
This paper presented a complete pipeline for semantic weed mapping using multispectral images and a DNN-based classifier. Our dataset consists of multispectral orthomosaic images covering 16,550\unit{m^2} sugar beet fields collected by five-band RedEdge-M and four-band Sequoia cameras in Rheinbach (Germany) and Eschikon (Switzerland). Since these images are too large to allocate on a modern GPU machine, we tiled them as the processable size of the DNN without losing their original resolution~of $\approx$1\unit{cm} GSD. These tiles are then input to the network sequentially, in a sliding window manner, for~crop/weed classification. We demonstrated that this approach allows for generating a complete field map that can be exploited for SSWM.

Through an extensive analysis of the DNN predictions, we obtained insight into classification performance with varying input channels and network hyperparameters. Our best model, trained on nine input channels (AUC of [\texttt{bg} = 0.839, \texttt{crop} = 0.863, \texttt{weed} = 0.782]), significantly outperforms a baseline SegNet architecture with only RGB input (AUC of [0.607, 0.681, 0.576]). In accordance with previous studies, we found that using the NDVI channel significantly helps in discriminating between crops and weeds by segmenting out vegetation in the input images. Simply increasing the size of the DNN training dataset, on the other hand, can introduce more ambiguous information, leading to lower~accuracy.

We also introduced spatiotemporal datasets containing high-resolution multispectral sugar beet/weed images with expert labeling.
Although the total covered area is relatively small, to our best knowledge, this is the largest multispectral aerial dataset for sugar beet/weed segmentation publicly available.
For supervised and data-driven approaches, such as pixel-level semantic classification, high-quality training datasets are essential. However, it is often challenging to manually annotate images without expert advice (e.g., from agronomists), details concerning the sensors used for data acquisition, and well-organized field sites. We hope our work can benefit relevant communities (remote sensing, agricultural robotics, computer vision, machine learning, and precision farming) and enable researchers to take advantage of a high-fidelity annotated dataset for future work. In our work, an~unresolved issue is limited segmentation performance for weeds in particular, caused by small sizes of plant instances and their natural variations in shape, size, and appearance. We hope that our work can serve as a benchmark tool for evaluating other crop/weed classifier variants to address the mentioned issues and provide further scientific contributions.

\vspace{6pt} 

\authorcontributions{I.S., R.K., P.L., F.L., and A.W. planned and sustained the field experiments; I.S., R.K., and P.L. performed the experiments; I.S. and Z.C. analyzed the data; P.L. and M.P. contributed reagents/materials/analysis tools; C.S., J.N., A.W., and R.S. provided valuable feedback and performed internal reviews. All authors contributed to writing and proofreading the paper.}
\funding{This project has received funding from the European Union's Horizon 2020 research and innovation programme under Grant No. 644227, and the Swiss State Secretariat for Education, Research and Innovation (SERI) under contract number 15.0029.}

\acknowledgments{We also gratefully acknowledge the support of NVIDIA Corporation with the donation of the Titan X Pascal GPUs used for this research, Hansueli Zellweger for the management of the experimental field at the ETH plant research station, and Joonho Lee for supporting manual annotation.}

\conflictsofinterest{The authors declare no conflict of interest.} 

\abbreviations{The following abbreviations are used in this manuscript:\\


\noindent 
\begin{tabular}{@{}ll}
RGB & red, green, and blue \\
UAV & Unmanned Aerial Vehicle \\
DCNN & Deep Convolutional Neural Network\\
CNN & Convolutional Neural Network\\
OBIA & Object-Based Image Analysis\\
RF & Random Forest\\
SSWM & Site-Specific Weed Management\\
DNN & Deep Neural Network\\
GPS & Global Positioning System \\
INS & Inertial Navigation System\\
DSM & Digital Surface Model \\
GCP & Ground Control Point \\
CIR &  Color-Infrared \\
NIR & Near-Infrared\\
NDVI & Normalized Difference Vegetation Index\\
GSD & Ground Sample Distance \\
FoV & Field of View \\
GPU & Graphics Processing Unit \\
AUC & Area Under the Curve \\
PA & Pixel Accuracy\\ 
MPA & Mean Pixel Accuracy \\
MIoU & Mean Intersection over Union\\
FWIoU & Frequency Weighted Intersection over Union\\
TF & True Positive\\
TN & True Negative\\
FP& False Positive\\
FN& False Negative\\
\end{tabular}}


\bibliographystyle{unsrt}
\reftitle{References}
\externalbibliography{yes}


\end{document}